\pdfoutput=1
\documentclass[11pt]{article}

\usepackage{EMNLP2023}

\usepackage{times}
\usepackage{latexsym}

\usepackage[T1]{fontenc}
\usepackage[utf8]{inputenc}

\usepackage{microtype}

\usepackage{inconsolata}

\usepackage{booktabs}       % professional-quality tables
\usepackage{amsfonts}       % blackboard math symbols
\usepackage{nicefrac}       % compact symbols for 1/2, etc.
\usepackage{microtype}      % microtypography
\usepackage{xcolor}         % colors
\usepackage{xspace}
\usepackage{amsmath}
\usepackage{optidef}
\usepackage{comment} % allow \begin{comment} \end{comment}
\usepackage{multirow}
\usepackage{amsthm}
\usepackage[normalem]{ulem}
\usepackage{wrapfig}
\newtheorem{definition}{Definition}

\usepackage{colortbl}
\usepackage{arydshln}
\usepackage[first=0,last=9]{lcg}
\usepackage{graphicx}
\usepackage{tablefootnote}

\newcommand\ximing[1]{{\color{purple}[{#1}]$_{-Ximing}$}}

\newcommand\qin[1]{{\color{red}[{#1}]$_{-Lianhui}$}}

\usepackage[linecolor=orange,size=scriptsize]{todonotes}

\newcommand{\method}{IPA\xspace}
\newcommand{\ipam}{$\text{IPA}^{\text{-}}$}
\newcommand{\realtoxic}{\textsc{RealToxicityPrompts}\xspace}
\newcommand{\csqa}{\textsc{CommonsenseQA}\xspace}

\newcommand{\ipa}[0]{IPA\xspace}
\newcommand{\ipastar}[0]{IPA*\xspace}

\newcommand*\inlineimage[1]{\raisebox{-0.15\baselineskip}{$\,$\includegraphics[height=0.81\baselineskip]{#1}$\,\,$}}
\newcommand{\beer}{\inlineimage{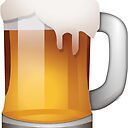}}

\newcommand*\inlinelargeimage[1]{\raisebox{-0.15\baselineskip}{$\,$\includegraphics[height=0.9\baselineskip]{#1}$\,\,$}}
\newcommand{\largebeer}{\inlinelargeimage{figures/beer.png}}

\title{\largebeer 
Inference-Time Policy Adapters (IPA):\\
Tailoring Extreme-Scale LMs without Fine-tuning
}

\newcommand{\aspace}{\hspace{1em}}
\newcommand{\uw}{$^{\heartsuit}$}
\newcommand{\aiTwo}{$^{\clubsuit}$}
\newcommand{\usc}{$^{\diamondsuit}$}

\author{
    Ximing Lu\uw \aiTwo \aspace 
    Faeze Brahman\uw \aiTwo \aspace 
    Peter West \uw \aiTwo \aspace
    Jaehun Jung \uw \aspace \aspace\\
    \textbf{Khyathi Chandu} \aiTwo \aspace  
    \textbf{Abhilasha Ravichander} \aiTwo \aspace 
    \textbf{Lianhui Qin} \uw \aspace \\ 
    \textbf{Prithviraj Ammanabrolu} \uw \aiTwo \aspace 
    \textbf{Liwei Jiang} \uw \aiTwo \aspace 
    \textbf{Sahana Ramnath} \usc \aspace \\
    \textbf{Nouha Dziri} \aiTwo \aspace
    \textbf{Jillian Fisher} \uw \aspace
    \textbf{Bill Yuchen Lin} \aiTwo \aspace
    \textbf{Skyler Hallinan} \uw \aspace\\
    \textbf{Xiang Ren} \usc \aiTwo \aspace
    \textbf{Sean Welleck} \uw\aiTwo \aspace 
    \textbf{Yejin Choi}\uw\aiTwo \aspace \\
    \aiTwo Allen Institute for Artificial Intelligence \\
    \uw   University of Washington \aspace
    \usc University of Southern California
}

\begin{document}
\maketitle
\begin{abstract}
%Large language models excel at a variety of language tasks when prompted with examples or instructions. 
%Yet controlling these models through prompting alone is limited.
%Tailoring language models through fine-tuning (e.g., via reinforcement learning) can be effective, but it is expensive 
%and requires model access.
While extreme-scale language models have demonstrated exceptional performance on a variety of language tasks, the degree of control over these language models through pure prompting can often be limited.
Directly fine-%\peter{removed 'fine-' because I think this is often more associated with supervised learning. But we can add it back}
tuning 
such language models can be effective for tailoring them, but it can be either extremely costly (e.g., GPT-3) or not even % available as an option
feasible for the broader community (e.g., GPT-4).

We propose \textbf{Inference-time Policy Adapters (IPA)}, which efficiently tailors a language model such as GPT-3 without fine-tuning it.
\ipa guides a large base model during 
decoding time through a lightweight policy adapter trained to optimize an arbitrary user objective with reinforcement learning. 

On five challenging text generation tasks, such as toxicity reduction and lexically constrained generation, \ipa consistently brings significant improvements over off-the-shelf language models.
It outperforms competitive baseline methods, sometimes even including expensive fine-tuning. 
In particular, tailoring GPT-2 with \ipa can outperform GPT-3, while tailoring GPT-3 with \ipa brings a major performance boost over GPT-3 (and sometimes even over GPT-4). 
%\ipa provides a lightweight way to 
%tailor large language models to a variety of objectives.
Our promising results highlight the potential of \ipa as a lightweight alternative to 
tailoring extreme-scale language models.\footnote{Our code is publicly available at: \url{https://github.com/GXimingLu/IPA}}
\end{abstract}
\section{Introduction}

Large language models (LLMs) have recently shown remarkable progress in various text generation tasks by adapting to instructions or examples~\citep{ouyang2022training, gpt-3}.
% (controllable)
%or a few in-context examples~\citep{gpt-3}.  
However, the degree of control (e.g., the inclusion of keywords, avoiding harmful language) offered by these extreme-scale models
%in zero/few-shot settings 
through pure prompting is still limited~\cite{Lou2023IsPA, Webson2021DoPM}. %\qin{We may give some examples for "control", so readers know what we mean by "control" specifically. e.g. "Lexically constrained generation is to ensure inclusion of keywords in output."} 
Moreover, prompting can be a brittle process due to LLMs being overly sensitive to the surface-form of the instructions \cite{perez2021true,lu-etal-2022-fantastically}. Furthermore, even with a carefully written prompt, LLMs may still struggle to fulfill certain task requirements due to their inherent limitations~\cite{liu2022tokenlevel, zong2022survey}.

Resource-intensive fine-tuning, through supervised learning, and more recently reinforcement learning (RL) \cite{DBLP:journals/corr/abs-2205-13636} have shown promise in tailoring language models to arbitrary user-given objectives. 
%However, they can be computationally expensive if not outright infeasible~\cite{OpenAI}. This issue is even more severe when access to the model's parameters (e.g., GPT-4) or powerful computational resources is restricted. 
%Fine-tuning through supervised learning, and more recently reinforcement learning (RL), has shown promise in tailoring language models to arbitrary objectives. 
RL, in particular, known for its generalizability and flexibility, allows models to learn from desired rewards.
However, these methods require accessing and updating models parameters, which can be extremely large or inaccessible in state-of-the-art models like GPT-4 \cite{OpenAI}. This limitation makes fine-tuning unfeasible for the broader community.

\begin{figure}
    \centering
    \includegraphics[clip,trim={0.40cm 0.40cm 0 0},
    width=0.49\textwidth]{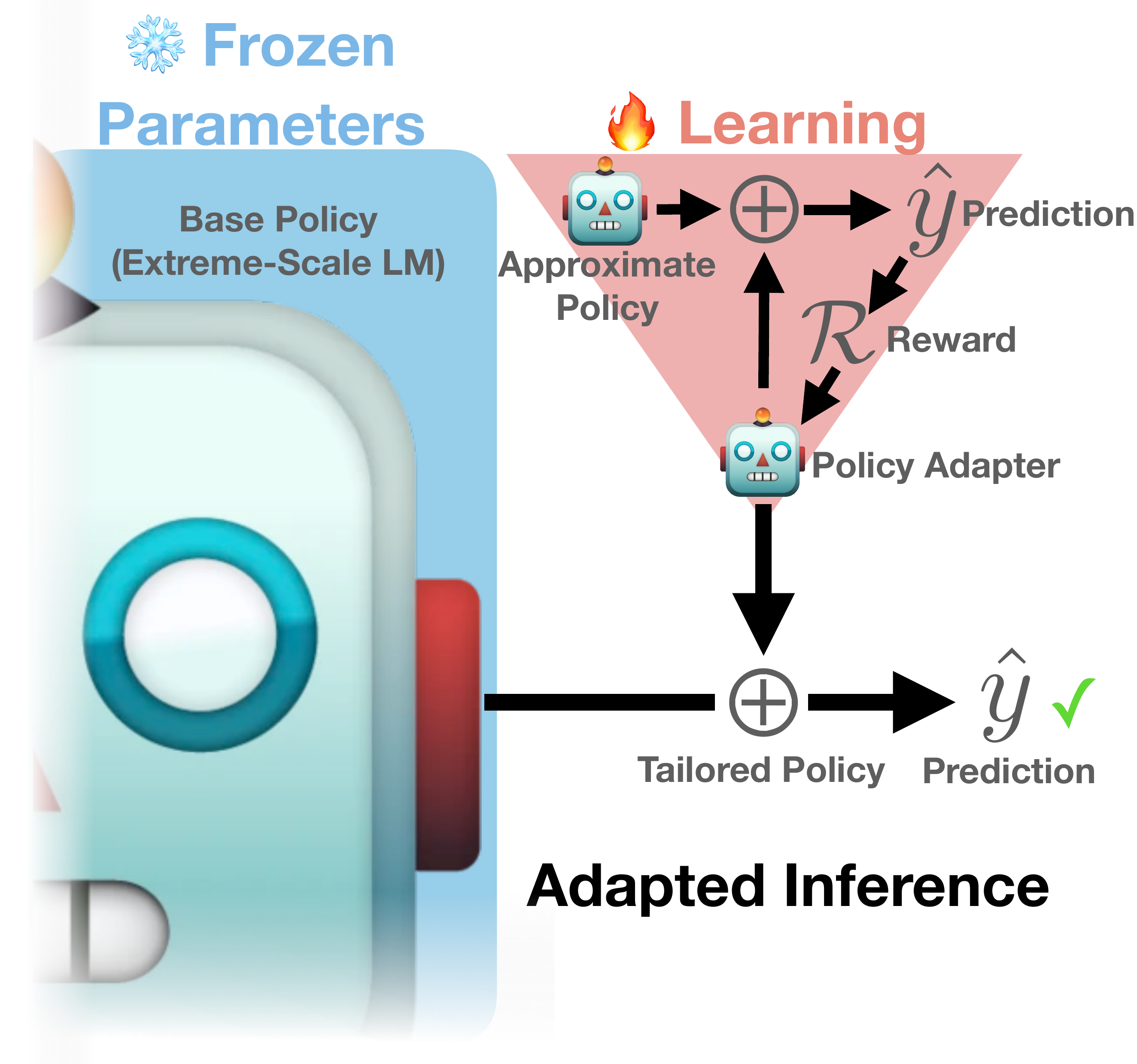}
    \caption{Inference-time Policy Adapters (IPA) efficiently steer a large-scale language model (such as GPT-3) during decoding-time through a lightweight policy adapter trained to optimize any arbitrary user objective with reinforcement learning. 
    %IPAs avoid the high cost of reinforcement learning by only sampling from the frozen base policy, or a more efficient approximate policy. % (above)\fb{what is above referring to?}. %\nouha{can we update the figure in a way that reflects the namings in the method section "base policy", "adaptor policy", "tailored policy"}\peter{will add this right now!}
    }
 %\vspace{-10pt}
    \label{fig:intro_fig}
\end{figure}

% Alternatively, \textit{inference}-time algorithms can tailor a language model  without accesssing to its parameters. 
% These algorithms adjust the model's output distribution using task-specific heuristics, while leaving the underlying model untouched. 
% However, 
% inference-time heuristics are traditionally hand-designed for specific tasks
% (e.g., \citet{DBLP:conf/naacl/LuWZBBC21, lu2020neurologic,DBLP:conf/acl/LiuSLSBSC20, Yang_2021,qin2022cold}),
% and these inference-time methods are often
% less effective than fine-tuning a model with RL \cite{DBLP:journals/corr/abs-2205-13636}.

%On the other hand, \textit{inference}-time algorithms offer an efficient alternative to tailor the output of LLMs. 
Alternatively, \textit{inference}-time algorithms can tailor a language model without accessing its parameters.
These algorithms align language models' outputs with desired task/user-specific properties 
%such as safety, logical coherence, faithfulness, etc, 
by adjusting the model's output distribution based on certain task-specific heuristics, while leaving the underlying model untouched. Despite the progress, these approaches are either restricted to specific tasks \cite{DBLP:conf/naacl/LuWZBBC21, lu2020neurologic}, require domain-specific knowledge \cite{DBLP:conf/acl/LiuSLSBSC20, Yang_2021}, suffer from expensive run-time at inference \cite{qin2022cold, qin2021future, dathathri2020plug}, or have shown to be less effective compared to direct RL optimization \cite{DBLP:journals/corr/abs-2205-13636}.

\begin{comment}
Alternatively, \textit{inference}-time algorithms can tailor a language model without accessing its parameters. 
These algorithms align language models' outputs with desired task/user-specific properties 
by adjusting the model's output distribution while keeping the underlying model untouched.
%They achieve this by adjusting the model's output distribution while keeping the underlying model untouched.
However, existing \textit{inference}-time algorithms often rely on hand-designed heuristics for specific tasks
(e.g., \citet{DBLP:conf/naacl/LuWZBBC21, lu2020neurologic,DBLP:conf/acl/LiuSLSBSC20, Yang_2021,qin2022cold}) and have shown to be less effective compared to direct RL optimization \cite{DBLP:journals/corr/abs-2205-13636} 
\qin{do we have more paper to cite?}.
\end{comment}

Drawing inspiration from RL and inference-time techniques,
we propose \textbf{Inference-time Policy Adapters (\beer IPA)}, an efficient and generalizable algorithm,
which tailors a large language model at decoding-time toward desired objectives without fine-tuning it. 
To do so, \ipa combines a large base LM's output distribution with that of a smaller-sized model (a lightweight \textbf{adapter policy}), and
optimizes the combined distribution towards a given objective with RL (Figure~\ref{fig:intro_fig}). 
\begin{comment}
During RL training, we keep the parameters of the target LM frozen and solely update the parameters of the adapter. 
%In situations where the target model is extremely expensive (such as GPT-3), we utilize a proxy model to approximate the target model's distribution for the RL training, which can either be a smaller model from the same language model family or a distilled version of the target model. 
In situations where the target model is extremely expensive (such as GPT-3), we utilize an \textbf{approximate policy}, a smaller proxy model that approximates the target model's distribution, during the RL training. The approximate policy can either be a smaller model from the same language model family or a distilled version of the target model.
At inference time, we decode with the combined distribution of the target model and the trained policy adapter.

Drawing inspiration from RL and inference-time techniques, 
we propose \textbf{Inference-time Policy Adapters (\beer IPA)}, 
which tailors a large language model at inference-time toward desired objectives without fine-tuning it. 
To do so, \ipa combines a large base LM's output distribution with that of a smaller-sized model (a lightweight \textbf{adapter policy}), and
optimizes the combined distribution towards a given objective with RL (Figure~\ref{fig:intro_fig}). 
\end{comment}
\ipa uses two key ideas to make learning efficient. First,
IPA \textit{only updates the adapter's parameters}, avoiding the need to update the large base LM. 
Second,
IPA replaces the large base model with an \textit{approximate policy}--a smaller model that approximates the base model's distribution.
The approximate policy is either a smaller model from the same language model family or a distilled version of the base model.
At inference time, we decode with the combined distribution of the base model and the trained policy adapter.

Experiments across five challenging text generation tasks show that \ipa brings consistent improvements over off-the-shelf language models, outperforming competitive baselines --- sometimes even including expensive fine-tuning.
In particular, tailoring GPT-2 with \ipa can outperform GPT-3, while tailoring GPT-3 with \ipa brings a major performance boost over GPT-3 (and sometimes even over GPT-4). 
Our compelling highlight the promise of \ipa as a lightweight alternative for tailoring large language models to a wide range of objectives. \ipa opens new ways to 
augment or customize extreme-scale language models using only academic-level resources.

\section{Background}
In this section, we introduce our text generation setting
(\S\ref{subsec:problem}) 
and a brief background on tailoring language models with reinforcement learning (\S\ref{subsec:RL}).
We then introduce our \ipa algorithm for tailoring large language models without fine-tuning (\S\ref{subsec:IPA}).
\subsection{Problem Setting}
\label{subsec:problem}
Text generation is the task of generating an output sequence $\textbf{y}$ given an input sequence $\textbf{x}$.
We consider standard autoregressive language models, which decompose a sequence's probability as $p_\theta(\textbf{y}|\textbf{x}) = \prod^{|\textbf{y}|}_{t=1} p_\theta(\textbf{y}_t|\textbf{y}_{<t}, \textbf{x})$, where $p_\theta$ is a neural network with parameters $\theta$.
Intuitively, our goal is to `tailor' a pretrained model $p_\theta$ towards a user-specified objective (e.g., safety).
Concretely, we assume that the objective is quantified by a reward function $\mathcal{R}(\textbf{y})\in \mathbb{R}$.
We then aim to adjust $p_\theta$ so that its generated sequences have high reward and reasonable language quality (e.g., fluency).

\subsection{Preliminary: Tailoring LMs with RL} \label{subsec:RL}
Online policy-based reinforcement learning has emerged as an effective way to adjust a language model towards a reward function.
Formally, these algorithms (e.g., PPO \cite{stiennon2022learning}, Quark \cite{Lu2022QuarkCT}, or NLPO \cite{Ramamurthy2022IsRL}) optimize a language model $p_\theta$ towards generating outputs $\textbf{y}$ that maximize a given reward $\mathcal{R}$:
% \vspace{-15pt}
\par\nobreak
\vspace{-6mm}
\begin{align*}
    \theta^\star = \arg\max \mathbb{E}_{\textbf{y}\sim p_\theta(\cdot | \textbf{x})}\mathcal{R}(\textbf{y}),
\end{align*}
\par\nobreak
\vspace{-1mm}
\noindent often along with regularization to maintain language quality.
% 
%At a high-level, these algorithms use a policy $p_\theta$ to collect input-output examples, score the outputs with a reward function $\mathcal{R}$, and produce an optimized policy $p_{\theta^\star}$. 
At a high-level, these algorithms use a policy $p_\theta$ to collect input-output examples, score the outputs with a reward function $\mathcal{R}$, and update parameter $\theta$ to maximize the expected reward. Although the exact optimization may differ, we can view
any online policy-based RL algorithms as a functions $f_{\textnormal{RL}}$ that take a policy $p_\theta$ and a reward function $\mathcal{R}$ as the inputs and outputs an optimized policy $p_{\theta^\star}$ with respect to $\mathcal{R}$. % via gradient ascent on $\theta$.
Formally,
\par\nobreak
\vspace{-6mm}
\begin{align} 
\label{eqn:rl}
% \mathcal{F}_{\textnormal{RL}} = \{ 
f_{\textnormal{RL}}: (p_\theta, \mathcal{R}; \theta') \rightarrow \theta^\star. 
% \}.
\end{align}
\par\nobreak
\vspace{-1mm}
\noindent Here $\theta'\subseteq\theta$ denotes the subset of $p_\theta$'s parameters that are updated by the algorithm.
The key idea behind \ipa is to use a full model $p_\theta$ to collect examples, but update a small set of parameters $\theta'$.

\section{Inference-time Policy Adapters (IPA)}

% \subsection{Tailoring LLMs with IPA} 
\label{subsec:IPA}
We introduce Inference-time Policy Adapters (IPA), a lightweight approach to tailor language models towards a user-specified objective. 
\ipa trains a small \textit{adapter policy} that adjusts the outputs of a (larger)  base model at inference-time in order to maximize a reward.
In doing so, \ipa avoids the cost of updating the large base model, without the need to hand-design inference-time heuristics.

\subsection{Policy Adaptation}
We introduce the notion of `\textit{tailoring}' used by \ipa, which mainly involves three policies. % using three policies.
First, \ipa starts with a \textbf{base policy} $p_\theta$, which is the language model to tailor.
Second, \ipa introduces an \textbf{adapter policy} $p_\phi$, which is a language model with the same output space as the base policy (i.e., vocabulary), but different parameters $\phi$.
Finally, \ipa combines the base and adapter policies into a \textbf{tailored policy}:
\begin{definition}[Tailored policy]
The tailored policy $p_{\theta \leftarrow \phi}$ combines the distributions of the base policy $p_{\theta}$ and the adapter policy $p_{\phi}$,
\vspace{-2.5mm}
\begin{align*} 
p_{\theta \leftarrow \phi} (\textbf{y}_{t}|\textbf{y}_{<t}) = \frac{1}{Z}  p_{\theta} (\textbf{y}_{t}|\textbf{y}_{<t}) p_{\phi}(\textbf{y}_{t}|\textbf{y}_{<t}),
\end{align*}
where $Z$ is a normalization factor.
\end{definition}
\noindent The tailored policy is a product-of-experts~\cite{10.1162/089976602760128018}, which amounts to multiplying the next-token probabilities from the base and adapter policies, then normalizing the result.
\ipa's tailored policy has two key properties. First, it allows for adjusting the base policy's output without direct access to the base policy's parameters. This is critical for tailoring modern LLMs that provide access to the model's output distribution but not the model's parameters.
Second, the policy adapter can use a much smaller model (i.e., $\phi \ll \theta)$.
This provides an efficient way to tailor a large base model.

\subsection{Adapter Training with RL}
Our goal is to adjust the tailored policy towards a user-specified objective.
The key idea in \ipa is to train the tailored policy to optimize a given reward with reinforcement learning, while \textit{only updating the parameters of the adapter policy}. 

Concretely, we use a reinforcement learning algorithm $f_{\textnormal{RL}}$ (Eqn.~\ref{eqn:rl}) to optimize
the tailored policy $p_{\theta \leftarrow \phi}$ with a reward function $\mathcal{R}$.
Notably, we keep the base policy's parameters ($\theta$) frozen, and only update the adapter policy's parameters ($\phi$).
That is,
\par\nobreak
\vspace{-6mm}
\begin{align*}
\phi^\star = f_{\textnormal{RL}} \: (p_{\theta \leftarrow \phi}, \mathcal{R}; \phi).
\end{align*}
\par\nobreak
\vspace{-1mm}
\noindent Intuitively, the adapter policy $p_\phi$ learns to rescale the frozen base policy $p_\theta$, yielding a tailored policy that is `tailored to' the reward.
Notice that our framework does not depend on a specific RL algorithm, but rather treats RL as a flexible plug-in optimization tool.
As we will demonstrate later, \ipa proves to be effective when paired with three different RL algorithms~\cite{Lu2022QuarkCT,Schulman2017ProximalPO, ramamurthy2023reinforcement}, and in principle, it can easily integrate with others.

\paragraph{Approximate Policy.} When the base model is extremely large (e.g., GPT-3), its forward pass is too costly to be used in the RL training loop. 
To overcome this, we propose using an  \textbf{approximate policy} in \ipa. %, which is a smaller model that approximates the base model's distribution.
\begin{definition}[Approximate policy]
The approximate policy is defined as a smaller-sized neural model parameterized by $\hat{\theta}$ that approximates the distribution of the base policy and is used to replace the base policy in the RL-based adapter training:
\vspace{-1mm}
\begin{align*}
\phi^\star = f_{\textnormal{RL}} \: (p_{\hat{\theta} \leftarrow \phi}, \mathcal{R}; \phi).
\end{align*}
\end{definition}
\vspace{-1mm}
\noindent In practice, we can obtain an approximate policy in two different ways. 
First, 
we can use a \textit{smaller pre-trained language model from the same model family}.
We do this if the smaller model has similar conditional generation behavior  
as the base policy.
For instance, we use an off-the-shelf GPT2-XL as the approximate policy to tailor GPT-3 in an open-ended generation. 
Alternatively, we can use a \textit{distilled base policy} as the approximate policy.
A distilled base policy is a language model trained on generations from the base policy, $\hat{\theta} = \arg\max \mathbb{E}_{\textbf{y} \sim p_{\theta}(\cdot | \textbf{x})}\big[\log P_{\hat{\theta}}(\textbf{y})\big]$, known as sequence-level knowledge distillation \cite{kim2016sequencelevel,west-etal-2022-symbolic}. For example, to tailor GPT-3 for lexically constrained generation, we tune GPT2-XL on prompt-generation pairs from GPT-3 to get a distilled base policy.

\paragraph{\ipa at Inference Time.} At inference time, \ipa uses the tailored policy $p_{\theta \leftarrow \phi}$ for decoding. 
Namely, at each time-step we obtain the next-token distribution from the tailored policy $p_{\theta \leftarrow \phi} (\textbf{y}_{t}|\textbf{y}_{<t})$, which can then be used with a standard decoding algorithm (e.g. nucleus sampling).

\section{Experiments}

%\fb{where did we explain IPA,IPA-,IPA* yet?}

We evaluate \method on a diverse range of tasks: toxicity reduction (\S\ref{section: detox}), lexically constrained generation (\S\ref{section: lexical}), open-ended generation (\S\ref{section: open_ended}), dialogue safety control (\S\ref{section: dia_safe}), and knowledge-grounded dialogue (\S\ref{section: dia_faith}). 
%, and self-rationalization in question-answering (\S\ref{section: rationale_acceptability}). 
In all benchmarks, \method consistently improve upon LLMs such as GPT-3 (\texttt{text-davinci-003}), surpassing competitive baselines and sometimes even outperforming expensive fine-tuned GPT-3 at a fraction of the cost.

% \jaehun{perhaps we could specify GPT-3/3.5 and GPT-4 briefly here, as readers maybe confused by specifically which model we are referring to with GPT-3/3.5/4}

%\ximing{TODO: add IPA, IPA-, IPA*}

\subsection{Toxicity Reduction}
\label{section: detox}

%Ensuring the safe deployment of language model is important yet challenging: due to misaligned pretraining objectives (i.e. modeling internet text vs. non-toxic text), 
LMs are susceptible to 
generating toxic completions, even when prompted with seemingly innocuous text~\cite{gehman-etal-2020-realtoxicityprompts}. Here, we assess IPA's efficacy in reducing toxicity from LMs.

%Here, we investigate the extent to which \ipa is effective in reducing the toxicity of language models in an open-ended setting. 
%. Even though recent models, like GPT-3, have implemented explicit toxicity guards through content filtering, they can still generate controversial or sensitive content when prompted with certain text. 
%\cite{Welleck2022GeneratingSB}. Additionally 
% Another practical application t

\paragraph{Datasets and Metrics.}
The task is to generate a fluent continuation $y$ while avoiding offensive content for a given prompt $x$. We evaluate this on \realtoxic benchmark \cite{gehman-etal-2020-realtoxicityprompts}, which contains 100k prompts designed to elicit toxic generations. 
%Following the experimental setup of \citep{liu-etal-2021-dexperts}, during training we use 85K prompts from the training set, and for evaluation we use the same 10K non-toxic prompts from test set as \citep{liu-etal-2021-dexperts}. 
Following the experimental setup of \citet{liu-etal-2021-dexperts}, we use Perspective API to  
determine the average maximum toxicity across 25 sampled generations and the (empirical) toxicity probability of at least one toxic generation.
% measure maximum toxicity, which is defined as the average maximum toxicity over 25 sampled generations, and the (empirical) toxicity probability of at least 1 out of 25 generations being toxic. 
In addition, we report fluency as the perplexity of generated output based on an off-the-shelf GPT2-XL model, and diversity as the count of unique n-grams normalized by the length of text. We also perform human evaluations; see Appendix \ref{tab:more_human} for more details.
%Finally, we conduct a pairwise human evaluation on toxicity, topicality, and fluency; more details are provided in Appendix \ref{human_detail} \fb{do we have human eval for this task?}.

\paragraph{Setup and Baselines}

%We use GPT-2 \fb{size?} as both the base and adapter policies.  
% We evaluate toxicity reduction with \method over off-the-shelf GPT-2 Large and GPT-3. For tailoring GPT-2, we use 
% Setup. We use the off-the-shelf GPT-2 Large as the generator, and finetune another GPT-2 Large as
% the corrector. 
We apply \method to tailor off-the-shelf GPT-2 and GPT-3\footnote{We refer \texttt{text-davinci-003} as GPT-3 in this paper}. To tailor GPT-2, we directly apply the base policy in the adapter training, denoted as IPA(GPT-2). For tailoring GPT-3, we use an off-the-shelf GPT-2 and a distilled GPT-3 \footnote{We finetune a GPT2-XL with 
prompt-output pairs from
GPT-3 on \realtoxic as the distilled GPT-3.} as the approximate policy for the adapter training, labeled as \ipam(GPT-3) and IPA*(GPT-3) respectively. Notice that \ipam(GPT-3) is equivalent to directly applying the policy adapter trained to tailor GPT-2 on top of GPT-3. 
%In all scenarios, we initialize the policy adapter with a pre-trained GPT2-large model.
We initialize all the policy adapters with a pre-trained GPT2-L model.

We use \textsc{Quark} as the RL algorithm in adapter optimization, and provide additional ablation studies %in Appendix \ref{tab:more_ablation} 
to assess the effects of different RL algorithms. 
We use the Perspective API as the reward function, which provides a score ranging from 0 to 1 to indicate the degree of toxicity.
%For more analysis on the reward function, please see Appendix \ref{tab:reward_analysis}.
%which measures the toxicity of the completed sequence. During inference, we use nucleus sampling with p = 0.9 to generate 25 samples for all baselines. 

For tailoring GPT-2, we compare \method with
% \ipa with its target LM, i.e., GPT-2, and
previously reported baselines from~\citet{DBLP:journals/corr/abs-2205-13636}, including decoding-based methods: PPLM~\citep{dathathri2020plug}, GeDi~\citep{krause-etal-2021-gedi-generative}, DExpert~\citep{DBLP:conf/acl/LiuSLSBSC20}, and learning-based methods: DAPT~\citep{gururangan-etal-2020-dont}, PPO~\citep{Schulman2017ProximalPO}, and \textsc{Quark}~\citep{DBLP:journals/corr/abs-2205-13636}. For tailoring GPT-3, we compare \method to the baselines described above that are compatible with GPT-3's limited accessibility: DExpert~\citep{DBLP:conf/acl/LiuSLSBSC20} and DAPT~\citep{gururangan-etal-2020-dont}. We also provide runtime analysis in Appendix \ref{tab:runtime}. 

%We use the Perspective API as a reward function, which provides a score between 1 (non-toxic) and 0 (toxic).

\paragraph{Results} 
\begin{table}[t]
\centering
\resizebox{.49\textwidth}{!}{
\begin{tabular}{lccccc}

\toprule
\multirow{2}{*}{\textbf{Models}} & \multicolumn{2}{c}{\textbf{Toxicity}} & \multicolumn{1}{c}{\textbf{Fluency}} & \multicolumn{2}{c}{\textbf{Diversity}} \\

\cmidrule(lr){2-3}\cmidrule(lr){4-4}\cmidrule(lr){5-6}

& \textbf{Avg Max.} & \textbf{Prob.} & \textbf{Pl.} & \textbf{Dist-2.} & \textbf{Dist-3.} \\
                          
\bottomrule

\rowcolor[gray]{0.90} \multicolumn{6}{l}{\textit{base policy:}  GPT2-L}\\
GPT-2 &  0.527 & 0.520 & 11.31 &	0.85 &	0.85 \\
% \bottomrule
PPLM &  0.520 &	0.518 & 32.58 &	0.86 &	0.86 \\
GeDi & 0.363 &	0.217 & 60.03 &	0.84 &	0.83 \\
\textsc{Dexperts} & 0.314 &	0.128 & 32.41 &	0.84 &	0.84 \\
DAPT & 0.428 &	0.360 & 31.21 &	0.84 &	0.84 \\
\midrule
PPO	& 0.218 &	0.044 & 14.27 &	0.80 &	0.84 \\
\textsc{Quark} & \underline{0.196} &	\underline{0.035} & \underline{12.47} &	0.80 &	0.84 \\
\midrule
IPA (GPT-2)	& \textbf{0.138} &	\textbf{0.031} & \textbf{11.94} &	0.80 &	0.84 \\

\bottomrule

\rowcolor[gray]{0.90} \multicolumn{6}{l}{\textit{base policy:} GPT-3} \\
 GPT-3 & 0.275 & 0.197 & 10.65 & 0.78 & 0.81 \\
% \bottomrule

\textsc{Dexperts} & 0.223 & 0.112 & 23.41 & 0.79 & 0.82 \\	
DAPT & 0.254 & 0.176 & 20.19 & 0.80 & 0.83 \\	
\midrule
\ipam$\>$ (GPT-3)	& 0.150 & 0.056 & \textbf{10.34} & 0.79 & 0.81 \\	
IPA* (GPT-3)	& \textbf{0.101} & \textbf{0.028} & 12.68 & 0.79 & 0.83 \\	

\bottomrule
\end{tabular}
}
\caption{Automatic evaluation for \textit{Toxicity Reduction} with off-the-shelf GPT2-large (top) and GPT-3 (bottom) as the base policy to tailor.}
\label{tab:toxicity}
\end{table}
\begin{table}[t]
\centering \footnotesize
\resizebox{.49\textwidth}{!}{
\begin{tabular}{lccccc}

\toprule
\multirow{2}{*}{\textbf{RL Algo.}} & \multicolumn{2}{c}{\textbf{Toxicity}} & \multicolumn{1}{c}{\textbf{Fluency}} & \multicolumn{2}{c}{\textbf{Diversity}} \\

\cmidrule(lr){2-3}\cmidrule(lr){4-4}\cmidrule(lr){5-6}

& \textbf{Avg Max.} & \textbf{Prob.} & \textbf{Pl.} & \textbf{Dist-2.} & \textbf{Dist-3.} \\
                          
\midrule
Quark	& 0.138 & 0.031 & 11.94 & 0.80 & 0.84 \\	
PPO	& 0.125 & 0.029 & 12.47 & 0.80 & 0.84 \\	
NLPO & 0.136 & 0.032 & 12.13 & 0.80 & 0.85 \\

\bottomrule
\end{tabular}
}
\caption{Comparison of using different RL algorithm for training IPA for \textit{Toxicity Reduction} with off-the-shelf GPT2-large as the base policy to tailor.}
\label{tab:toxicity_ablation}
\end{table}

\definecolor{grey}{HTML}{7f7f7f}
\definecolor{blue}{HTML}{0000D1}
\begin{figure}[t]
    \centering
    \includegraphics[width=0.44\textwidth]{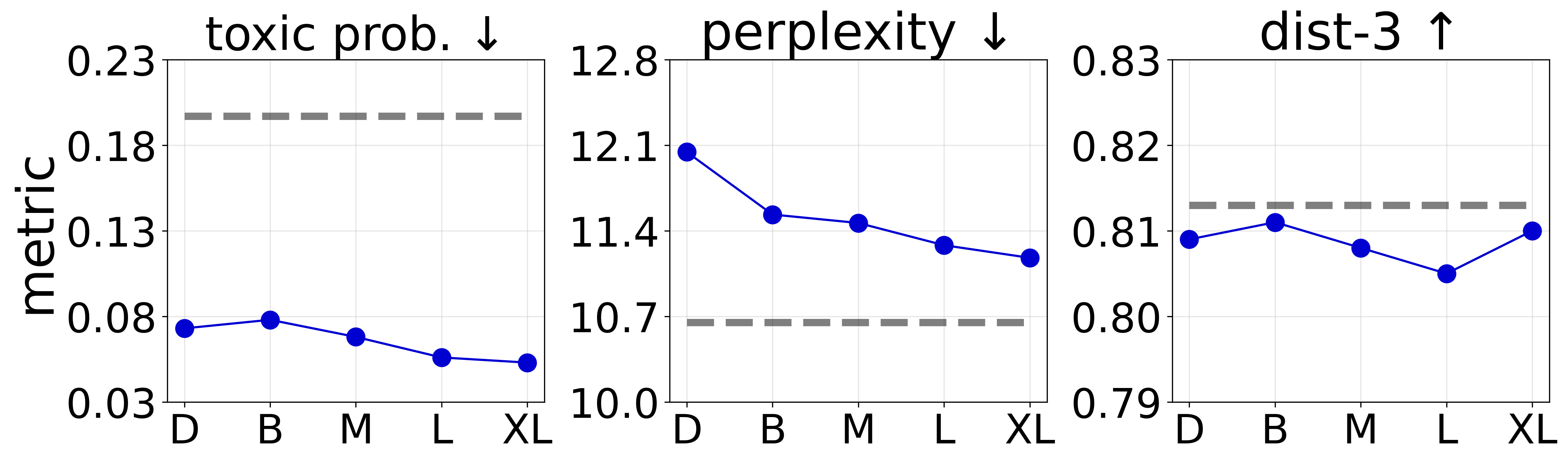}
    \caption{Performance of \ipam\phantom{.}(\textbf{\textcolor{blue}{blue line}}) with respect to the size of the adapter model (distill-GPT2, GPT2-small, GPT2-medium, GPT2-large, GPT2-XL) on top of a off-the-shelf GPT-3 as the base policy. The \textbf{\textcolor{grey}{grey line}} denotes the performance of the off-the-shelf GPT-3. 
    }
    \label{fig:policy_size}
\end{figure}

As shown in Table \ref{tab:toxicity}, \ipa outperforms all learning-based and decoding-based methods in tailoring GPT-2 and GPT-3, significantly reduces the toxicity while maintaining language quality. Interestingly, we found that applying the policy adapter optimized for GPT-2 directly on top of GPT-3 (i.e., \ipam) is highly effective, showcasing the adaptability and reusability of \method. 
Notably, when tailoring GPT-3, IPA outperforms the costly domain adaptive training (DAPT), which exhaustively fine-tune GPT-3 on a non-toxic corpus. This further emphasizes the promise of the \method as a cost-efficient approach to align LLMs. Our findings are further confirmed by human evaluation (Appendix \ref{tab:more_human}).

Finally, we conduct ablations on the effect of RL algorithms. As shown in Table \ref{tab:toxicity_ablation}, IPA is effective with various RL algorithms, all of which lead to state-of-the-art performance. Additional ablation in Figure \ref{fig:policy_size} shows that a policy adapter as small as a distilled GPT-2 can effectively tailor the $\times1000$ larger GPT-3 model, achieving comparable performance with our main result.

%Finally, we conduct additional ablations in Appendix \ref{tab:reward_analysis}, and demonstrate that 1) IPA is effective with various RL algorithms, all of which lead to strong performance; and 2) a policy adapter as small as a distilled GPT-2 can effectively tailor the $\times1000$ larger GPT-3, achieving similar performance to our main result. 

%Finally, as shown in Table \ref{tab:toxicity_ablation}, %ablation studies on using different RL algorithms for optimizing the policy adapter show that 
%\method can smoothly accommodate different RL algorithms, all leading to better performance compared to other baselines.

\subsection{Lexically Constrained Generation}
\label{section: lexical}

Next, we test IPA in lexically constrained generation.
%, where given a set of constraint words, the model needs to generate a sentence that includes all the given constraints. While prior works on this task evaluated the order-invariant satisfaction of the constraints -- i.e. the generation was deemed to be correct when it simply includes all the keywords \cite{commongen, lu2020neurologic}, 
We consider a more challenging setup of \textit{ordered lexical constraints}, where the generation is considered correct if it includes all the keywords with the correct order specified in the input prompt.

\paragraph{Datasets and Metrics.}
We use \textsc{CommonGen} \cite{commongen}, a dataset for generative commonsense reasoning.
%where the task is to generate a coherent sentence given a set of concept words. 
%To evaluate both the inclusion and order of the given constraints, 
We deliberately instruct the models to generate a sentence with the given keywords while following the order they appear in the input prompt. For automatic evaluation, we gauge the constraint satisfaction with \textit{coverage}, a binary metric that evaluates a generation to be correct only when it includes all the keywords and also matches the specified order. We also measure the \textit{fluency} 
%of each generation 
using a critic model fine-tuned on CoLA \cite{cola}. For human evaluation, we assess the \textit{quality} and \textit{plausibility} of model generations for 100 randomly sampled test examples based on a 3-point Likert Scale; see details in Appendix \ref{human_detail}.

\begin{table}[t]
\centering
\resizebox{.43\textwidth}{!}{
\begin{tabular}{lccccc}

\toprule
\multirow{2}{*}{\textbf{Models}} & \multicolumn{2}{c}{\textbf{Automatic}} & \multicolumn{3}{c}{\textbf{Human}} \\
\cmidrule(lr){2-3}\cmidrule(lr){4-6}

& \textbf{Cov.} & \textbf{Fl.} & \textbf{Qu.} & \textbf{Pl.} & \textbf{Overall} \\
                          
\midrule
                
%GPT-3$_\text{distill}$ & 31.93 & 90.11 & 2.76 & 2.72 & 2.60 \\
GPT-3 & 37.01 & 94.89 & 2.84 & 2.81 & 2.60\\
GPT-3.5 & 65.17 & \textbf{95.89} & \underline{2.93} & \underline{2.88} & \underline{2.90} \\
GPT-4 & 84.81 & \underline{95.49} & \textbf{2.95} & \textbf{2.97} & \textbf{2.96} \\

\midrule

GPT-3$_\text{sft}$ & 72.89 & 73.96 & 2.56 & 2.60 & 2.50 \\

\midrule

%\ipa (GPT-3$_\text{distill}$) & \underline{85.68} & 91.67 & 2.70 & 2.64 & 2.53 \\
$\text{\ipa}^*$ (GPT-3) & \textbf{88.54} & 92.58 & 2.90 & 2.87 & 2.88\\

\bottomrule
\end{tabular}
}
\caption{Automatic and human evaluation results for \textit{Lexically Constrained Generation}. Human evaluation scores are on a 3-point Likert Scale.\tablefootnote{Human pairwise agreements are 0.97, 0.94, and 0.93 for quality, plausibility and overall, respectively.}
} 
\label{tab:commongen}
\end{table}

\paragraph{Setup and Baselines.}
As we will demonstrate later, zero-shot GPT-3 is surprisingly poor at satisfying ordered lexical constraints, even with explicit instructions. Our goal is to make GPT-3 more reliable in constraint satisfaction.
We use distilled GPT3 \footnote{We finetune a GPT2-XL with prompt-output pairs from GPT-3 on \textsc{CommonGen} train set as the distilled GPT-3} as the approximate policy for adapter training, since an off-the-shelf GPT-2 cannot perform constrained generation out of the box. We initialize the policy adapter with a pre-trained GPT2-L model.
%As off-the-shelf GPT-2 cannot reliably perform lexically constrained generation, we use GPT-2-XL fine-tuned with GPT-3 generations on CommonGen train set as our base policy (GPT-3$_\text{distill}$) and GPT-2-Large as our adapter policy.
We use \textsc{Quark} as the RL algorithm and
choose our reward to be the product of the coverage score and the fluency score, as this promotes constraint satisfaction and fluency preservation. Please see Appendix \ref{tab:reward_analysis} for more reward analysis.

We compare IPA with its base policy GPT-3, as well as more advanced LLMs: GPT-3.5 and GPT-4 \cite{gpt-4}. As a strong supervised baseline, we also fine-tune GPT-3 on the \textsc{CommonGen} train set, which contains human-written outputs with the correct lexical order, denoted as GPT-3$_\text{sft}$.

\paragraph{Results.}
As shown in Table \ref{tab:commongen}, powerful LMs such as GPT-3 often struggle to satisfy ordered lexical constraints even with explicit instructions.
%We present the experimental results in Table \ref{tab:commongen}. Surprisingly, powerful language models such as GPT-3 struggle at faithfully following ordered lexical constraints even with explicit instructions. 
\method leads to remarkable improvement on top of GPT-3 and surpasses more advanced models such as GPT-3.5 and GPT-4 in terms of constraint coverage, while achieving better or comparable generation quality. Noticeably, \method outperforms fine-tuned GPT-3 in both constraint coverage and generation quality at a fraction of its cost: while fine-tuning GPT-3 costs \$156.82, training a distilled GPT-3 as the approximate policy requires only \$28.59 for generating outputs from GPT-3. Our results highlight the potential of the \method as a cost-efficient way to enhance the capabilities of LLMs.

\begin{table}[t]
\centering
\footnotesize
\renewcommand{\arraystretch}{1.25}
\setlength\tabcolsep{2.5pt}
\resizebox{.43\textwidth}{!}{
\begin{tabular}{lcccc}
\hline
\textbf{Decoding Method} & \textbf{Diversity} & \textbf{Coherence} & \textbf{Critic} & \textbf{Mauve} \\ \hline
\rowcolor[gray]{0.90} \multicolumn{5}{l}{\textit{base policy}: GPT2-XL} \\
greedy                                                               & 55.05                                      & \underline{49.57}                                & 7.88                                    & 15.32          \\
top-k (k=50)                                                            & 92.60                                      & 48.53                                      & 10.72                                   & 53.13          \\
top-p (p=0.95)                                                          & 95.85                                      & 47.61                                      & 13.24                                   & 56.42          \\
typical ($\tau$=0.95)                                                         & 95.80                                      & 46.08                                      & 23.49                                   & \underline{63.92}    \\
SimCTG                                                               & 95.67                                      & 46.12                                      & 23.67                                   & 62.21          \\
Contrastive                                                          & \underline{95.99}                                & 49.42                                      & \underline{36.73}                             & 61.95          \\ \hline % \cdashline{1-5}
\ipa (GPT2-XL)                                                                & \textbf{96.12}                             & \textbf{51.81}                             & \textbf{50.93}                          & \textbf{84.18} \\ \hline
\rowcolor[gray]{0.90} \multicolumn{5}{l}{\textit{base policy}: GPT-3} \\ 
top-p (p=0.95) & \underline{95.63} & 56.16 & 18.58 & 63.73 \\ \hline %\cdashline{1-5}
\ipam $\>$ (GPT-3) & 95.35 & \underline{57.26} & \underline{22.62} & \underline{71.40} \\
\ipastar(GPT-3) & \textbf{96.26} & \textbf{61.94} & \textbf{32.84} & \textbf{73.17} \\ \hline
\end{tabular}
}
\caption{Automatic evaluation for \textit{open-domain generations} on XSum with off-the-shelf GPT2-XL (top) and GPT-3 (bottom) as the base policy to tailor. Critic scores refer to \textit{human-likeness} according to OpenAI detector.}
\label{tab:open-domain-auto}
% \vspace{-5mm}
\end{table}
\subsection{Open-ended generation}
\label{section: open_ended}
We further evaluate \ipa on open-ended generation,
%To evaluate the effectiveness of \ipa in a general setting, we conduct experiments on an open-ended generation task; 
following the experimental setup in \cite{DBLP:journals/corr/abs-2210-15097}. The goal is to make machine-generated content more fluent, coherent, and human-like.

\paragraph{Datasets and Metrics.} We experiment on the news domain using XSum dataset \cite{narayan-etal-2018-dont}. Following \citet{DBLP:journals/corr/abs-2210-15097}, 
%we filter out news articles with fewer than 160 tokens. 
we use the first 32 words as our input prompt, and generate 84 tokens as continuations. We evaluate using both automatic and pairwise human evaluation. For automatic evaluation, we use aggregate n-gram diversity and coherence scores \cite{DBLP:journals/corr/abs-2210-15097} as well as MAUVE \cite{pillutla-etal:mauve:neurips2021}, which measures the distribution similarity between the set of human-written and machine-generated texts. To measure the \textit{human-likeness} of generated texts, we employ OpenAI detector\footnote{\small{\url{https://github.com/promptslab/openai-detector}}}, a classifier for distinguishing AI vs. human-written text. We use the classifier's probability assigned to `human' text to serve 
as an additional metric, denoted as Critic.
\noindent For human evaluation, we randomly sample 100 test examples and 
%generate continuations using different models. We 
perform pairwise comparisons of our method against baselines on \textit{coherence} and \textit{fluency} using AMT; see details in Appendix \ref{human_detail}.

\paragraph{Setup and Baselines.}

We apply \method to tailor off-the-shelf GPT2-XL and GPT-3, following the same setup as toxicity reduction task (section \ref{section: detox}). Same as before, the tailor policies are denoted as IPA(GPT-2), \ipam(GPT-3) and IPA*(GPT-3), respectively. We use \textsc{Quark} as the RL algorithm and the product of diversity, coherence, and critic scores as the reward function. We found it critical to combine multiple metrics as the reward function to improve the overall generation quality;  see Appendix \ref{tab:reward_analysis} for more analysis on reward functions.

%To tailor GPT-2, we directly apply the base policy in the adapter training, referred to as IPA(GPT-2). For tailoring GPT-3, we use an off-the-shelf GPT2-XL and a distilled GPT-3 as the approximate policy for the adapter training, denoted as IPA-(GPT-3) and IPA*(GPT-3) respectively. Notice that IPA-(GPT-3) is equivalent to directly applying the policy adapter trained to tailor GPT-2 on top of GPT-3. In all these scenarios,
%we initialize the policy adapter with a pre-trained GPT2-large model, and use \textsc{Quark} as the RL algorithm to optimize the policy adapter. As the reward function, we use the multiplication of diversity, coherence, and critic scores described above. 

\begin{figure}
    \centering
    \includegraphics[
    width=0.45\textwidth]{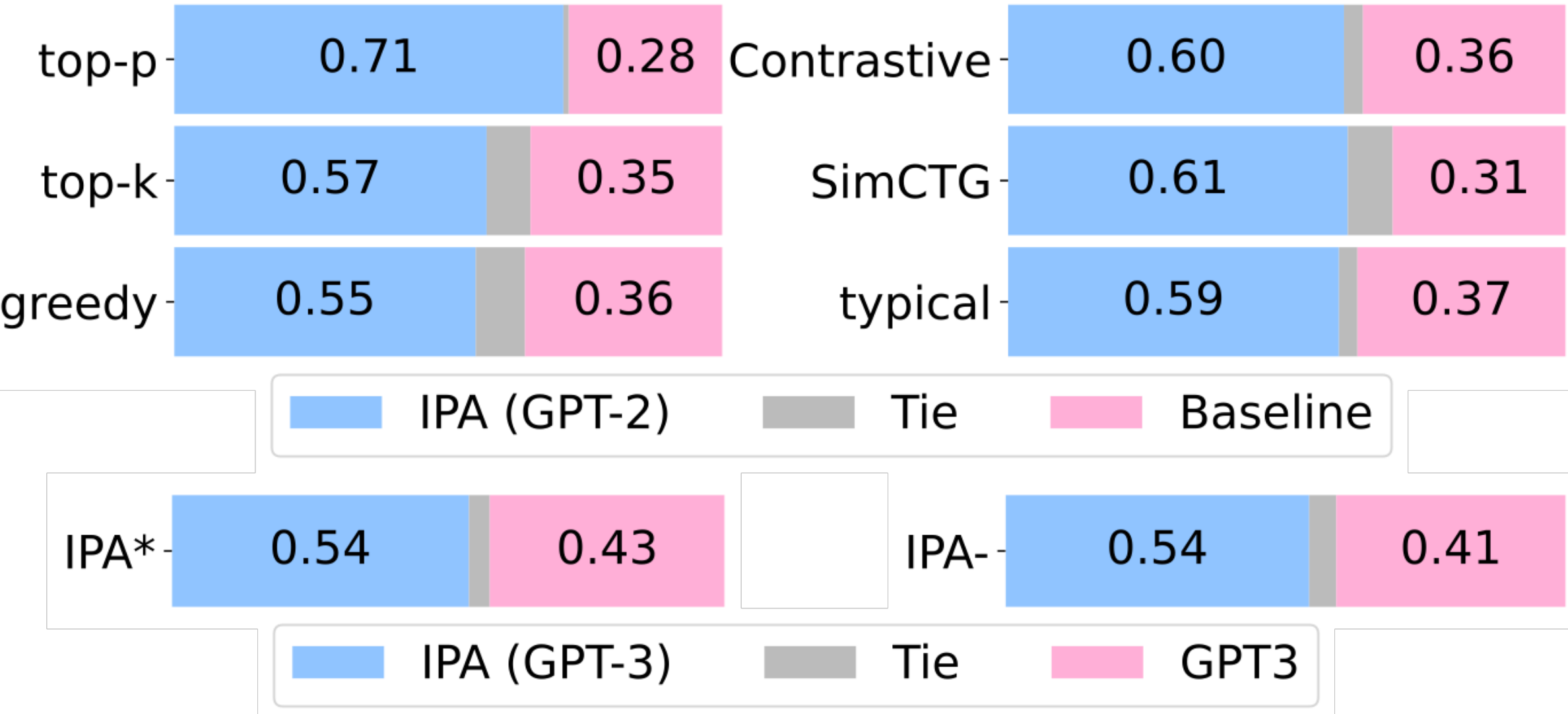}
    \caption{Pairwise human evaluation in terms of \textbf{overall quality} for \textit{Open-ended Generation} on XSum with off-the-shelf GPT2-XL (top) and GPT-3 (bottom) as the base policy to tailor.\tablefootnote{Average pairwise agreements are 0.88 and 0.82 with GPT2-XL and GPT-3, respectively.}
    }
    \vspace{-10pt}
    \label{fig:open_human}
\end{figure}

\noindent For tailoring GPT-2, we compare decoding with \ipa with six different decoding strategies: greedy, top-k sampling ($k=50$), nucleus sampling ($p=0.95$), typical sampling ($\tau=0.95$) \cite{typical}, SimCTG \cite{NEURIPS2022_871cae8f}, and Contrastive decoding \cite{DBLP:journals/corr/abs-2210-15097}. The latter three are specifically designed to improve the coherence and naturalness of the generated text.
%For all decoding baselines, we use the recommended hyper-parameters in the papers.
For tailoring GPT-3, we compare \ipa with GPT-3's default generation technique: decoding with nucleus sampling ($p=0.95$). as other decoding methods are not applicable to GPT-3 due to its limited API access.
%through API interface.

%\noindent Additionally, we include comparisons with GPT-3 when using GPT2-XL as the proxy policy \ipa- (GPT-3), and distilled GPT-3 as the proxy policy \ipastar(GPT-3).

\paragraph{Results.} As shown in Table \ref{tab:open-domain-auto}, \ipa significantly outperforms all previous baselines in tailoring GPT-2 and GPT-3 across all automatic metrics. Notably, it achieves an absolute improvement of 20.26\% over the best-performing baseline in the Mauve score.
% which has shown a high correlation with the human judgment. %of quality. 
Our pairwise human evaluation in Figure \ref{fig:open_human} also verify the results. \ipa generates significantly more coherent and fluent texts compared to other baselines. Overall, on average, human evaluators preferred \ipa 1.8$\times$ more than other baselines. 
%The lower section of Table \ref{tab:open-domain-human} shows that \ipastar applied on top of GPT-3 is preferred over its base policy GPT-3 on fluency, coherence, and overall quality.
Interestingly, we found that directly applying the policy adapter optimized for GPT-2 on top of GPT-3 (i.e., \ipam) significantly improves the generation quality, highlighting the adaptability and reusability of \method. 
We observed further improvement when using distilled GPT-3 as the approximate policy (i.e., IPA*). Our promising results once again showcase the effectiveness and efficiency of \ipa.

% \fb{additional human results added/todo: update this}

%\input{tables/results_open_domain_human}

\subsection{Dialogue Safety Control}
\label{section: dia_safe}

% \begin{table}[t]
% \centering
% \footnotesize
% \renewcommand{\arraystretch}{1.5}
% \setlength\tabcolsep{2.7pt}
% \begin{tabular}{lcccccc}

% \midrule
% \multirow{2}{*}{\textbf{Models}}   & \multicolumn{3}{c}{\textbf{Automatic}}                       & \multicolumn{1}{l}{\textbf{}} & \multicolumn{2}{c}{\textbf{Human}}   \\ \cline{2-4} \cline{6-7} 
%                           & \textbf{Safety} & \textbf{Coh.} & \textbf{Engag.} & \textbf{}                     & \textbf{Safety} & \textbf{Coh.} \\ \hline
% DialoGPT         & 0.46            &                    &                       &                               & 1.34            & 2.45               \\
% Godel           & 0.49            &                    &                       &                               & 1.40            & 2.53               \\
% Blenderbot-3B       & 0.53            & 0.82               & 1.50                  &                               & 1.43            & 2.60               \\
% ChatGPT         & 0.74            &                    &                       &                               & \textbf{1.60}   & 2.68               \\ \hline
% IPA- (BlenderBot-3B) & \textbf{0.78}   & \textbf{0.88}      & \textbf{2.68}         &                               & \underline{1.57}      & \textbf{2.75}      \\ \midrule
% \end{tabular}
% \caption{Automatic and human evaluation results for \textit{Dialogue Safety Control}. Human evaluation scores are on a 3-point Likert Scale.}
% \label{tab:dialog_safe}
% \end{table}

%%% version wo/ automatic coherence and engagingeness

\begin{table}[t]
\centering
\footnotesize
\renewcommand{\arraystretch}{1.25}
\setlength\tabcolsep{2.7pt}
\resizebox{.42\textwidth}{!}{
\begin{tabular}{lcccc}
\midrule
\multirow{2}{*}{\textbf{Models}} & \textbf{Automatic} & \multicolumn{1}{l}{\textbf{}} & \multicolumn{2}{c}{\textbf{Human}}   \\ \cline{2-2} \cline{4-5} 
                                 & \textbf{Safety}    & \textbf{}                     & \textbf{Safety} & \textbf{Coherence} \\ \hline
DialoGPT              & 0.46               &                               & 1.34            & 2.45               \\
Godel                   & 0.49               &                               & 1.40            & 2.53               \\
Blenderbot              & 0.53               &                               & 1.43            & 2.60               \\
ChatGPT                 & 0.74               &                               & \textbf{1.60}   & 2.68               \\ \hline
\ipam $\>$(BlenderBot-3B)        & \textbf{0.78}      &                               & \underline{1.57}      & \textbf{2.75}      \\ \midrule
\end{tabular}
}
\caption{Automatic and human evaluation results for \textit{Dialogue Safety Control}. Human evaluation scores are on a 3-point Likert Scale.\tablefootnote{Human pairwise agreements are 0.84 and 0.87 for safety and coherence.}
}
% \vspace{-5pt}
\label{tab:dialog_safe}
\end{table}
Existing dialogue systems often fail to respond safely to potentially unsafe user utterances \cite{kim-etal-2022-prosocialdialog}, limiting their deployment in real-world applications. Here, we aim to evaluate \ipa for controlling the safety of a dialogue model.

\paragraph{Datasets and Metrics.} We experiment on \textsc{DiaSafety} \cite{sun-etal-2022-safety}, a challenging dataset containing ~54K context-sensitive unsafe examples. The task is to generate a coherent response to a potentially unsafe utterance while avoiding offensive, harmful, toxic or biased language. \textsc{DiaSafety} contains human-written safe and unsafe responses which we use to train a dialogue safety classifier. We use the classifier score as an automatic measure of safety. In addition, we conduct a human evaluation of \textit{safety} and \textit{coherence} (3-point Likert scale) on 200 examples through Amazon Mechanical Turk; see Appendix \ref{human_detail} Figure \ref{fig:mturk_dialogue_safety} for details. 

% perform safety evaluation using a classifier trained on ...
\paragraph{Setup and Baselines.} We apply \method to tailor the Blenderbot family models~\cite{roller-etal-2021-recipes}, which are pretrained dialogue agents. Specifically, we use Blenderbot-3B-distill as the frozen base policy, a samller Blenderbot-1B-distill as the approximate policy and initialize the policy adapter with a Blenderbot-1B-distill model. %\footnote{BlenderBot checkpoints are available at: \url{https://huggingface.co/models}} 
We use \textsc{Quark} as the RL algorithm for adapter training.
To preserve the dialogue quality while controlling the response safety, we choose our reward to be the product of the safety score from our dialogue safety classifier, as well as coherence and engagingness scores from UniEval-Dialogue \cite{zhong-etal-2022-towards}.\footnote{\url{https://github.com/maszhongming/UniEval}}
%During inference, we use Nucleus sampling \cite{Holtzman2020The} with $p=0.6$ and temperature 1.0.

\noindent We compare \ipa with its base policy, i.e., Blenderbot-3B-distill, and other off-the-shelf dialogue models including DialoGPT \cite{zhang-etal-2020-dialogpt}, GODEL \cite{peng2022godel} as well as ChatGPT \cite{team2022chatgpt}. ChatGPT is known to have safeguards through content filtering and is considered a strong baseline.

\paragraph{Results.} As shown in Table \ref{tab:dialog_safe}, \ipa significantly improves dialogue safety and coherence compared to
its base policy Blenderbot-3B-distill, surpassing other dialogue models including  DialoGPT and GODEL. In comparison with ChatGPT, \ipa achieves comparable performance on safety based on both automatic and human evaluation while showcasing improved coherence. Upon further investigation, we found that ChatGPT often generates canned responses like "I'm a language model; I'm not allowed..." as hard safeguards, which hurts the coherence and naturalness of the dialogue flow. On the other hand, Blenderbot tailored by \method can generate safe responses that are coherent, natural, and human-like. Our results demonstrate the potential of \method to enhance controllability in various NLP applications beyond conditional text generation. 
\subsection{Knowledge-grounded Dialogue}
\label{section: dia_faith}
Ideally, knowledge-grounded dialogue systems should generate responses that are faithful to the given knowledge. However, models tend to generate hallucination containing unverifiable information \cite{dziri2022faithdial, rashkin2021measuring, dziri2022evaluating}. To address this undesirable behavior, we use \ipa to tailor dialogue model towards generating more faithful content.  Given the knowledge $K$ and the conversation history $H$, the task is to generate a response $r$ that's faithful to $K$ and coherent with $H$. 

\paragraph{Dataset and Metrics}
We evaluate on the Wizard of Wikipedia (WoW) data. WoW \cite{dinanwizard} involves a Wizard and an Apprentice engaging in a conversation. The Wizard's role is to provide information on a specific topic, while the Apprentice's task is to seek further details. 
% During the conversation, the Wizard is presented with excerpts from Wikipedia and selects a relevant portion—usually one or two sentences—as evidence to support their response.
WoW has been shown to suffer from hallucinations \cite{dziri2022origin}, in more than 60\% of the turns, making it a valuable dataset for studying  hallucination issues. 
FaithDial \cite{dziri2022faithdial} is a hallucination-free benchmark created by modifying the hallucinated responses within the WoW dataset. 
We use the FaithDial test data at test time to evaluate the faithfulness of responses and compare them against the knowledge snippets and gold responses.   

To measure faithfulness, we use the critic model \cite{dziri2022faithdial}, which returns the 
probability of an given utterance being identified as faithful.
%percentage of s identified as faithful. 
Additionally, we use BERTScore to measure the semantic similarity between the generated response $r$ and the knowledge $K$, and the token-level F1 score to rate the lexical overlap between $r$ and $K$. To measure coherence and engagingness, we use the UniEval model \cite{zhong-etal-2022-towards}. %See \S\ref{} for more details.
\begin{table}[t]
\centering
\footnotesize
\renewcommand{\arraystretch}{1.25}
\setlength\tabcolsep{2pt}
\resizebox{.49\textwidth}{!}{
\begin{tabular}{lccccc}
\hline
\textbf{Dialogue Model} & \textbf{Critic} & \textbf{BERTScore} & \textbf{F1} & \textbf{Coherence} & \textbf{Engaging} \\ \hline
\rowcolor[gray]{0.90} \multicolumn{6}{l}{\textit{supervised} baseline} \\ 
GPT-2 & 39.9 & 0.29 & 47.7 & 0.77 & 1.26\\
\textsc{DialoGPT} & 40.6 & 0.34 & 53.5 & 0.83 & 1.32 \\
\textsc{DOHA} & 46.8 & 0.32 & 56.1 & 0.88 & \underline{1.33}\\
  \textsc{T5} & 53.5   & 0.41  &  61.7 &  0.86 &  1.28 \\
\textsc{T5-CTRL} & 54.8 & \underline{0.45} & \underline{65.2} & 0.83 & 1.21 \\ 
\textsc{T5-LT} & \underline{58.6}  & 0.43 & 65.0 & 0.83 & 1.21 \\
 \hline
\rowcolor[gray]{0.90} \multicolumn{6}{l}{\textit{off-the-shelf} dialogue model} \\ 

BlenderBot & 10.3  & 0.12 & 9.8 & \textbf{0.92} & 1.21 \\
 \midrule
\ipam $\>$(BlenderBot) &  \textbf{76.6}  & \textbf{0.68} & \textbf{80.1}  & \underline{0.91} & \textbf{1.34} \\
 \hline
\end{tabular}
}
\caption{Evaluation results for \textit{Knowledge-Grouded Dialogue} generations on Faithdial. We use off-the-shelf Blenderbot as the base policy to tailor.}
\vspace{-5pt}
\label{tab:faith-auto}
\end{table}

\paragraph{Setup and Baselines} Similar to the dialogue safety experiment, we use the Blenderbot-\{3, 1\}B-distill model \cite{roller-etal-2021-recipes} as our base policy and approximate policy respectively, and initialize the policy adapter with a Blenderbot-1B-distill model.
We use \textsc{Quark} as the RL algorithm. %. to optimize the policy adapter.
To preserve coherence and engagingness while ensuring the faithfulness of a dialogue response, we choose our reward to be the product of the faithfulness score from the critic model described above, as well as coherence and engagingness scores from UniEval-Dialogue \cite{zhong-etal-2022-towards}.
%\footnote{\url{https://github.com/maszhongming/UniEval}}
%During inference, we use Nucleus sampling \cite{Holtzman2020The} with $p=0.6$ and temperature 1.0.

We compare  to previously baselines from~\citet{dziri2022faithdial}, supervised models fine-tuned on WoW, including GPT2, DialoGPT \cite{zhang-etal-2020-dialogpt}, DoHA \cite{prabhumoye-etal-2021-focused} T5 \cite{raffel2020exploring}, T5-CTRL \cite{rashkin-etal-2021-increasing}, and T5-LossTruncation \cite{kang-hashimoto-2020-improved}. We also compare against the base policy, off-the-shelf BlenderBot model \cite{roller-etal-2021-recipes}.

%During inference, we use Nuclues sampling \cite{Holtzman2020The} with $p=0.6$ and temperature 1.0. One challenge that arises with knowledge-grounded dialogue systems is the generation of responses that are both faithful and highly extractive, simply by copying and pasting exact information from the knowledge source. Ideally, we aim for responses that strike a balance between faithfulness and abstractiveness. This means generating responses that go beyond mere copy-pasting of information and instead, demonstrate a creative use of the knowledge at hand. 
%\nouha{Mentioning this depends on the fact that we measure abstractiveness at test time}
%To ensure both faithfulness and abstractiveness, we employ a scoring mechanism that combines the faithfulness score generated by the critic model with an extractiveness score. The extractiveness score is determined by the density metric, which represents the average length of text spans copied from the knowledge that is included in the response. By multiplying these two scores together, we can effectively consider both the faithfulness of the response and the extent to which it relies on direct extraction from the knowledge source.

\paragraph{Results} As shown in Table \ref{tab:faith-auto}, supervised models struggle to generate faithful dialogue response grounded on the given knowledge. %even after being fine-tuned on hundreds of thousands of human-written dialogues. 
This is mainly because of the poor data quality of their supervision dataset: WoW has been shown to suffer from hallucinations in more than 60\% of the turns~\cite{dziri2022faithdial}.
Moreover, pre-trained dialogue models like Blenderbot demonstrate even worse performance at generating faithful response, despite being trained on WoW and other knowledge-grounded dialogue datasets in their pre-training stage. \method significantly improves the faithfulness of the generated dialogue response over its base policy Blenderbot while preserving the dialogue quality (i.e., coherence and engagingness), outperforming all other baselines. Our results showcases the potential of \ipa to improve reliability and trustworthiness in various 
 downstream applications.

\section{Related Work}
% \ximing{New outline}\\
%There are three major trends for controlling LMs: 

% \paragraph{Decoding Method}
%\cite{chen2022controllable},
\paragraph{Controlled Decoding}

Recent studies have explored controlled generation at inference time by designing new decoding algorithms \cite{keskar2019ctrl,mireshghallah2022mix, li2022diffusion, chen2022controllable, zhang2022survey}. For example, %to ensure the inclusion of given keywords, 
Neurologic decoding \cite{lu2020neurologic}, and GBS \cite{hokamp2017lexically} generalize beam search for lexically constrained decoding,  by constraining  decoding space with keyword-related penalties. DExperts \cite{liu-etal-2021-dexperts} modifies output distribution during decoding with attribute-specific expert models. %However, these decoding methods are designed for particular control types only. 
Another line of research develops gradient-based decoding for more general control \cite{qin2020back, qin2022cold, sha2020gradient, DBLP:conf/iclr/DathathriMLHFMY20,kumar2021controlled}. For example, COLD Decoding \cite{qin2022cold} introduces energy-based modeling to impose arbitrary constraints on text and samples with Langevin dynamics. %PPLM \cite{DBLP:conf/iclr/DathathriMLHFMY20} steers generation towards desirable attributes via the gradients from a small fine-tuned model. 
Despite their progress, these approaches either are designed for particular control types or rely on computationally expensive gradient computations. 

\paragraph{Reinforcement Learning for NLG}
RL has historically been used in multiple NLG tasks such as machine translation~\citep{wu2016google, nguyen-etal-2017-reinforcement}, 
summarization~\citep{paulus2017deep}, dialogue~\citep{li-etal-2016-deep, zhou2017endtoend}, text games~\citep{karthik2015language,jericho}, etc to
%ensure that the generated text is 
optimize for an arbitrary non-differentiable reward.
This was often done using online policy gradient methods such as REINFORCE~\citep{sutton2018reinforcement}, leading to documented issues with reward hacking%---where a model has great metrics but fails to solve the spirit of the task and produces inarticulate sounding text
~\citep{Choshen2020On,samuel2021revisiting}.
Recent advances %use the success of LMs in modeling language fluency to 
introduce a KL reward penalty which significantly increases the naturalness of generated text~\citep{ouyang2022training, korbak-etal-2022-rl}.
This method has been used extensively to tune a base LM via online on-policy~\citep{Ramamurthy2022IsRL}, off-policy~\citep{guo-etal-2022-efficient,Lu2022QuarkCT}, and offline~\citep{snell2023offline,korbak2023pretraining} RL.
Such methods quickly become computationally infeasible for extreme-scale LMs. %with billions of parameters 
\section{Conclusion}

we present \method, a lightweight inference-time policy adapter that tailor a frozen large language model towards desirable properties (e.g., safety, coherence) in an efficient,
generalizable, and flexible way. Specifically,
%we present \method, an efficient Inference-time Policy Adapter that tailor a frozen large language model towards desirable properties (e.g., safety, coherence, faithfulness). Specifically, 
%complements the undesirable properties (e.g., toxicity, incoherency, hallucination) of a frozen base policy
%in an efficient, generalizable, and flexible way. % and transferrable 
% \method inherits the generalizability of the RL approach to enable tailoring models with arbitrary objectives while incorporating plug-and-play flexibility of the inference-time techniques.
\method combines the generalizability of RL with the plug-and-play flexibility of inference-time techniques, permitting customization of large language models 
%for various tasks, 
without the need for costly fine-tuning. 
% \method can effectively customize the base policy models to specified desirable objectives without sacrificing the existing favorable attributes of the base models. 
% Crucially, \method enables tailoring massive-size large language models that are generally untunable due to computational cost or restricted API access. 
% As \method operates independently of the choice of the online learning RL algorithm, it is flexible to accommodate future alternative optimization techniques.
%Extensive experiments across five challenging text generation tasks demonstrate that \ipa consistently outperforms competitive inference-time baselines, occasionally surpassing even those enhanced by expensive supervision. This work aims to inspire efficient solutions that augment model scale pursuits using academic resources.
Extensive experiments across five challenging text generation tasks show that IPA brings consistent improvements over LLMs, outperforming competitive baselines — sometimes even surpassing expensive fine-tuning. 
We hope our work sheds light on creative and efficient algorithmic innovations to complement the pursuit of model scales with academic-level resources.
% In sum, \method presents a compelling and creative case of leveraging academic-level resources to seek efficient %alternative 
% algorithmic solutions to complement the pursuit of model scales.

%\newpage
%\clearpage

\section{Limitations and Ethical Consideration}

% , such as language safety, faithfulness, or constrained satisfaction.

While the versatility of the IPA is a crucial feature that enables aligning large language models with arbitrary user-given objectives, it may also pose potential dual-use concerns, especially when combined with the power of large language models.

%While the versatility of the IPA is a crucial feature that enables the flexlarge language models to be customized according to any user-defined objective, it may also pose potential dual-use concerns, especially when combined with the power of large language models
%In this work, we present \method, an Inference-time Policy Adapter module that can be flexibly applied to the output space of extreme-size large language models during inference time to steer their generations. IPA is trained with reinforcement learning guided by reward functions reflecting arbitrary objectives. Such flexibility of IPA, while enabling desirable capabilities of LLMs, may also pose potential dual-use concerns.

First, as with any controllable text generation technique, \method could be potentially used for unintended malicious purposes, such as manipulating models to produce hateful, toxic content or misinformation. As malicious users can already exploit any existing techniques for harmful purposes theoretically, we foresee minimal risk introduced by \method specifically. Nevertheless, we highly recommend avoiding such negative applications of \ipa. %In particular, as we only plan to release \method optimized for positive intentions, such as toxicity reduction, and dialogue safety control, we inherently promote reducing undesirable model behaviors to the best of our effort.

Moreover, similar to any RL-based method that depends on the reward function for learning signals, \method is susceptible to the innate shortcomings from the reward model. For instance, we use the Perspective API calls as the reward function for the toxicity reduction task; any limitations or potential biases from these public API calls will propagate into the learning of \method.
Nonetheless, as more accurate, transparent, and inclusive classifiers are developed, we anticipate that \method would inherit those improvements as well.

%The potential danger of such error propagation issue may be further exacerbated if the reward models themselves are highly opaque, and hence their downsides may only reveal if their signal has been amplified with downstream modules like \method. As \method is a highly efficient method for adapting reward signal compared to usual RL methods that require fine-tuning of the full model, we expect flexible and easy adaption of \method to future iterations of better reward models.

Beyond these two primary concerns, another inherent limitation of \method is its requirement to access the output logits of the base LM. 
This constraint hinders IPA's compatibility with certain models, such as GPT-4, which permit access only to the output, not the logits.
%This constraint restricts IPA's use with some models like GPT-4, which allow access only to the output, not the logits. 
Finally, like general RL frameworks, IPA relies on the assumption that user objectives are quantifiable through a reward function. However, this premise may not always hold, particularly when user objectives are inherently challenging to measure, thus limiting IPA's applicability.

\paragraph{}

\bibliography{anthology,custom}
\bibliographystyle{acl_natbib}
\paragraph{}
\newpage 
\pagebreak
\appendix
%\appendix

% \section{Further Experiments}
% \label{more_exp}
% \subsection{Additional ablations}
% \label{more_ablation}
% \subsection{Reward Analysis}
% \label{reward_analysis}

% \section{Runtime}
% \label{run_time}

%\section{Experimental Details}
\section{Further Experiment}
\subsection{Human Evaluation for Toxicity}
\label{tab:more_human}
We perform additional pairwise human evaluation on tailoring GPT-3 to reduce toxicity. We compare the outputs from IPA* and IPA- to each baseline, based on the perceived level of toxicity (which one is less rude or disrespectful), topicality (which one is more natural, relevant, and logical), and fluency (which one is more grammatically correct and coherent), on 100 random prompts from the test set of \realtoxic using.  

\newcolumntype{x}[1]{%
>{\centering\hspace{0pt}}p{#1}}%

\begin{table}[t]
\setlength{\tabcolsep}{2.4pt}
    \centering\footnotesize
        \scalebox{.82}{
    \begin{tabular}{l|x{11mm}p{8mm}|x{13mm}p{8mm}|x{11mm}p{8mm}}
        \toprule
        & \multicolumn{2}{c|}{\textbf{IPA- vs. GPT3}}  & \multicolumn{2}{c|}{\textbf{IPA- vs. D\tiny{EXPERTS}}} & \multicolumn{2}{c}{\textbf{IPA- vs. DAPT}} \\ \midrule
 %       \rowcolor[gray]{0.95} \multicolumn{7}{l}{\textit{Our Method: IPA-}} \\
     \textbf{Less Toxic}   & \textbf{0.17} & 0.09 & \textbf{0.15} & 0.09 & \textbf{0.13} & 0.12 \\
     \textbf{More Topical} \hspace{2mm} & 0.20 & \textbf{0.21} & \textbf{0.23} & 0.14 & \textbf{0.22} & 0.20\\
      \textbf{More Fluent}  & \textbf{0.27} & 0.23 & \textbf{0.24} & 0.16 &\textbf{0.21} & 0.18 \\
     %\bottomrule
    \end{tabular}}
    \newline
    \vspace{0mm}
    \newline
            \scalebox{.82}{
    \begin{tabular}{l|x{11mm}p{8mm}|x{13mm}p{8mm}|x{11mm}p{8mm}}
        \toprule
        & \multicolumn{2}{c|}{\textbf{IPA* vs. GPT3}}  & \multicolumn{2}{c|}{\textbf{IPA* vs. D\tiny{EXPERTS}}} & \multicolumn{2}{c}{\textbf{IPA* vs. DAPT}} \\ \midrule
 %       \rowcolor[gray]{0.95} \multicolumn{7}{l}{\textit{Our Method: IPA-}} \\
     \textbf{Less Toxic}   & \textbf{0.18} & 0.05 & \textbf{0.14} & 0.06 & \textbf{0.15} & 0.10 \\
     \textbf{More Topical} \hspace{2mm} & 0.23 & 0.23 & \textbf{0.28} & 0.17 & 0.18 & 0.18\\
      \textbf{More Fluent}  & \textbf{0.26} & 0.21 & \textbf{0.32} & 0.15 &\textbf{0.23} & 0.22 \\
     \bottomrule
    \end{tabular}}
    
    \caption{Human evaluation results of \textit{Toxicity Reduction}, comparing the percentage of texts rated as less toxic, more topical, and more fluent as generated by IPA- and IPA* versus other baselines.}
    \label{tab:toxicity_human_eval}
\end{table}

%For each prompt, we compare 6 pairs of models: IPA* versus other baselines and IPA- versus other baselines, as shown in Table X. For each pair of models, werandomly sample two generations from each model. In total we have 1200 comparisons, and each comparison is rated by 3 raters. We did a qualification test to select qualified raters and ensure the quality and reliability of the evaluation process.
As shown in Table \ref{tab:toxicity_human_eval}, the human evaluation results confirms that both IPA- and IPA* effectively tailor GPT-3 to be less toxic while maintaining the language quality. This again underscores the potential of IPA as a cost-effective method for aligning large language models with user-defined objectives.

\subsection{Additional Baseline: Few-shot}
\label{tab:few_shot}
In the experimental section, we show that in zero-shot setting LLMs such as GPT-3 often struggle to fulfill users' requests, such as generating safe content or reliably satisfying lexical constraints. Here, we conduct additional experiment to access LM's performance in few-shot setting on toxicity reduction and lexically constrained generation.

\begin{table}[t]
\centering \footnotesize
\resizebox{.49\textwidth}{!}{
\begin{tabular}{lccccc}

\toprule
\multirow{2}{*}{\textbf{Models}} & \multicolumn{2}{c}{\textbf{Toxicity}} & \multicolumn{1}{c}{\textbf{Fluency}} & \multicolumn{2}{c}{\textbf{Diversity}} \\

\cmidrule(lr){2-3}\cmidrule(lr){4-4}\cmidrule(lr){5-6}

& \textbf{Avg Max.} & \textbf{Prob.} & \textbf{Pl.} & \textbf{Dist-2.} & \textbf{Dist-3.} \\
                          
\bottomrule

% \rowcolor[gray]{0.90} \multicolumn{6}{l}{\textit{GPT-3}} \\
 GPT-3 (zero-shot) & 0.275 & 0.197 & 10.65 & 0.78 & 0.81 \\
% \bottomrule

GPT-3 (5-shot) & 0.214 & 0.132 & 15.96 & 0.76 & 0.80 \\	
GPT-3 (10-shot) & 0.208 & 0.145 & 17.83 & 0.77 & 0.80 \\	
\midrule
IPA- (GPT3)	& 0.150 & 0.056 & \textbf{10.34} & 0.79 & 0.81 \\	
IPA* (GPT3)	& \textbf{0.101} & \textbf{0.028} & 12.68 & 0.79 & 0.83 \\	

\bottomrule
\end{tabular}
}
\caption{Automatic evaluation results for \textit{Toxicity Reduction} with off-the-shelf GPT-3.}
\label{tab:toxicity_few_shot}
\end{table}

\begin{table}[t]
\centering \footnotesize
\resizebox{.36\textwidth}{!}{
\begin{tabular}{lcc}

\toprule

\textbf{Models} & \textbf{Coverage} & \textbf{Fluency} \\
                          
\midrule
                
%GPT-3$_\text{distill}$ & 31.93 & 90.11 & 2.76 & 2.72 & 2.60 \\
GPT-3 (zero-shot) & 37.01 & \textbf{94.89} \\
GPT-3 (5-shot) & 43.85 & \underline{94.34}  \\
GPT-3 (10-shot) & 45.70 & 94.21 \\

\midrule

%\ipa (GPT-3$_\text{distill}$) & \underline{85.68} & 91.67 & 2.70 & 2.64 & 2.53 \\
$\text{\ipa}^*$ (GPT-3) & \textbf{88.54} & 92.58 \\

\bottomrule
\end{tabular}
}
\caption{Automatic evaluation results for \textit{Lexically Constrained Generation} with off-the-shelf GPT-3.}
\label{tab:commongen_few_shot}
\end{table}

As illustrated in Table \ref{tab:toxicity_few_shot} and Table \ref{tab:commongen_few_shot}, prompting GPT-3 with additional few-shot examples improves its performance to some extent, but it still falls short of consistently fulfill users' requests. The gain is particularly limited in lexically constrained generation, likely due to GPT-3's inherent limitations when dealing with hard logical constraints. Importantly, IPA on top of zero-shot GPT-3 outperforms all the few-shot baselines by a noticeable margin across all scenarios. The results further highlight the importance of our method, which directly optimize the base policy to align with user-specified objectives instead of solely relying on the innate capabilities of LLMs through prompting.

\subsection{Additional Experiments with LLaMA}
We conducted additional experiments with LLaMA models \cite{Touvron2023LLaMAOA} for the constrained generation task. We apply IPA to tailor an off-the-shelf LLaMA-13B model and initialize the policy adapter with a LLaMA-7B model. As shown in Table \ref{tab:llama_lexical}, IPA leads to remarkable improvement on top of LLaMA-13B in terms of constraint coverage while maintaining language quality.

\begin{table}[t]
\centering \footnotesize
\resizebox{.34\textwidth}{!}{
\begin{tabular}{lcc}

\toprule

\textbf{Models} & \textbf{Coverage} & \textbf{Fluency} \\
                          
\midrule
                
%GPT-3$_\text{distill}$ & 31.93 & 90.11 & 2.76 & 2.72 & 2.60 \\
LLaMA	& 28.73	& 89.64 \\
\ipam$\>$ (LLaMA) &	\textbf{81.49} &	\textbf{89.71} \\
%coverage, fluency & 88.54 & \textbf{92.58}  \\
\bottomrule
\end{tabular}
}
\caption{Automatic evaluation results for \textit{Lexically Constrained Generation} with off-the-shelf LLaMA-13B as the base policy to tailor.}
\label{tab:llama_lexical}
\end{table}

\subsection{Reward Analysis}
\label{tab:reward_analysis}
\begin{table}[t]
\centering \footnotesize
\resizebox{.36\textwidth}{!}{
\begin{tabular}{lcc}

\toprule

\textbf{Reward} & \textbf{Coverage} & \textbf{Fluency} \\
                          
\midrule
                
%GPT-3$_\text{distill}$ & 31.93 & 90.11 & 2.76 & 2.72 & 2.60 \\
coverage & \textbf{90.75} & 83.91 \\
coverage, fluency & 88.54 & \textbf{92.58}  \\
\bottomrule
\end{tabular}
}
\caption{Automatic evaluation results for \textit{Lexically Constrained Generation} with off-the-shelf GPT-3 as the base policy using different reward functions}
\label{tab:reward_lexical}
\end{table}

\begin{table}[t]
\centering
\footnotesize
\renewcommand{\arraystretch}{1.25}
\setlength\tabcolsep{2.5pt}
\resizebox{.48\textwidth}{!}{
\begin{tabular}{lcccc}
\hline
\textbf{Reward} & \textbf{Diversity} & \textbf{Coherence} & \textbf{Critic} & \textbf{Mauve} \\ \hline
coherence                                                               & 92.41                                      & \textbf{64.98}                                & 5.41                                    & 68.25          \\
coherence, critic                                                       & \underline{93.73}                                & 51.03                                      & \textbf{52.36}                             & \textbf{84.32}          \\ 
coherence, critic, diversity                                                             & \textbf{96.12}                             & \underline{51.81}                             & \underline{50.93}                          & \underline{84.18} \\ \hline
\end{tabular}
}
\caption{Automatic evaluation for \textit{open-domain generations} on XSum with off-the-shelf GPT2-XL as the base policy using different reward functions.}
\label{tab:reward_open}
\end{table}

\begin{table}[t]
\centering
\footnotesize
\renewcommand{\arraystretch}{1.25}
\setlength\tabcolsep{2pt}
\resizebox{.49\textwidth}{!}{
\begin{tabular}{lcccc}
\hline
\textbf{Reward} & \textbf{Safety}  & \textbf{Coherence} & \textbf{Engaging} & \textbf{Overall}\\ \hline
safety & \textbf{0.85} & 0.82 & 1.32 & 0.88 \\
safety, coherence, engaging &  0.78  & \textbf{0.90} & \textbf{1.91} & \textbf{0.98}\\
 \hline
\end{tabular}
}
\caption{Evaluation results for \textit{Dialogue Safety Control} on \textsc{DiaSafety} with different reward functions.}
\vspace{-5pt}
\label{tab:reward_safe}
\end{table}

\begin{table}[t]
\centering
\footnotesize
\renewcommand{\arraystretch}{1.25}
\setlength\tabcolsep{2pt}
\resizebox{.49\textwidth}{!}{
\begin{tabular}{lcccc}
\hline
\textbf{Reward} & \textbf{Critic}  & \textbf{Coherence} & \textbf{Engaging} & \textbf{Overall}\\ \hline
critic & \textbf{85.3} & 0.84 & 1.01 & 0.88 \\
critic, coherence, engaging &  76.6  & \textbf{0.91} & \textbf{1.34} & \textbf{0.97}\\
 \hline
\end{tabular}
}
\caption{Evaluation results for \textit{Knowledge-Grouded Dialogue} on Faithdial with different reward functions.}
\vspace{-5pt}
\label{tab:reward_faith}
\end{table}
We provide further analysis to justify our selection of reward functions for each task.
\paragraph{Toxicity Reduction} Following previous work \citet{Lu2022QuarkCT}, we use the Perspective API score as a reward function, which provides a score between 1 (non-toxic) and 0 (toxic). We observe that IPA effectively reduce the toxicity while preserving the language quality in terms of fluency and diversity in both automatic and human evaluation.

\paragraph{Lexically Constrained Generation} Our goal is to enhance constraint satisfaction. As shown in Table \ref{tab:reward_lexical}, optimizing for constraint coverage alone may result in a slight decline in language fluency, as measured by COLA. However, by incorporating fluency as an auxiliary reward, we notice improvements in both dimensions. Human evaluations further support our findings.

\paragraph{Open-ended Generation} The goal is to make machine-generated content more fluent, coherent, and human-like. 
As shown in Table \ref{tab:reward_open}, optimizing solely for coherence does not yield significant improvements in the overall generation quality, as evaluated by MAUVE. Incorporating scores from the OpenAI detector, a classifier for distinguishing between AI vs. human-written text, as an additional reward serves as an essential element in improving the overall quality and human-likeness of generated texts. Moreover, we found that integrating diversity score as another auxiliary reward helps maintain the diversity of generations while promoting higher quality output.

\paragraph{Dialogue Safety Control} Our aim to improving the safety of a dialogue model. 
As shown in Table \ref{tab:reward_safe}, optimizing for safety score alone may result in a decrease in the overall quality of the generated dialogue, measured by coherence, engagingness and overall score from UniEval-Dialogue \cite{zhong-etal-2022-towards}. The generated responses tends to be bland and templated, such as "I don't know...", "I'm not sure...". 
We found that integrating coherence and engagingness scores as additional reward helps preserving natural dialogue flow while promoting safe responses. 

\paragraph{Knowledge-grounded Dialogue} Our aim to improving the faithfulness of dialogue response with respect to the given knowledge. 
As shown in Table \ref{tab:reward_faith}, optimizing for faithfulness score alone may result in a decrease in the overall quality of the generated dialogue, measured by coherence, engagingness and overall score from UniEval-Dialogue \cite{zhong-etal-2022-towards}. The generated responses are often the exact copy of the given knowledge, lacking of abstractiveness. 
We found that integrating coherence and engagingness scores as additional reward helps preserving the naturalness of the generated responses while enhancing their faithfulness.

\section{Runtime Analysis}
We conduction additional runtime analysis on toxicity reduction task, comparing the inference speed of IPA with other baseline methods. As shown in Table \ref{tab:runtime}, IPA is significantly more efficient than most of the baseline methods and falls within a similar range as nucleus sampling.
\label{tab:runtime}
\begin{table}[h]
\renewcommand{\arraystretch}{1.1}
\setlength{\tabcolsep}{13.6pt}
    \centering\footnotesize
      \resizebox{\linewidth}{!}{%
    \begin{tabular}{l | c}
     \specialrule{\heavyrulewidth}{-\heavyrulewidth}{2pt}
        \hspace{16mm}\textbf{Method} & \textbf{Runtime}  \\ \specialrule{\lightrulewidth}{-\lightrulewidth}{0pt}
        Nucleus Sampling & 0.03 \\
        PPLM \cite{dathathri2020plug} & 23.7 \\
        GeDi \cite {krause-etal-2021-gedi-generative}  & 0.78\\
        Dexperts \cite {DBLP:conf/acl/LiuSLSBSC20} & 0.12 \\
        DAPT \cite{gururangan-etal-2020-dont} & 0.03 \\
        \textsc{Quark} \cite{DBLP:journals/corr/abs-2205-13636} & 0.03 \\
        Inference-time Policy adapter & 0.08 \\
     \specialrule{\heavyrulewidth}{-\heavyrulewidth}{-1.5pt}
    \end{tabular}}
    \caption{Inference runtime (seconds per sentence generation) of IPA versus other baseline methods with GPT2-L as the base policy on toxicity reduction task.}
\end{table}

\section{Experiment Detail}
\subsection{Off-the-Shelf Models}
We download off-the-shelf models, including pretrained GPT-2 and BlenderBot, from HuggingFace Transformers \cite{wolf-etal-2020-transformers}, which are implemented in the PyTorch deep learning framework. We access GPT-3, GPT-3.5 and GPT-4 models via API calls through OpenAI platform.
\subsection{Model Training Details}
%\ximing{working on it..}
All training is performed on 8 NVIDIA Quadro RTX 8000 GPUs and costs about 3000 GPU hours in total. Our method is implemented with PyTorch an the Huggingface Transformers library.
\subsubsection{Toxicity Reduction}
We initialize the policy adapter with an off-the-shelf GPT2-L model and use \textsc{Quark} as the RL algorithm for the adapter training.  Hyperparameters for training are given in Table \ref{tab:hyper_toxic}. We performed a hyperparameter grid search for the number of training steps over the range [10k, 20k], for the KL coefficient $\beta$ over the range [0, 0.3], and for the frequency of exploration over the range [5, 20]. During inference, we use nucleus sampling with $p= 0.9$ and temperature 1.0.

\begin{table}[h]
\renewcommand{\arraystretch}{1.1}
\setlength{\tabcolsep}{13.6pt}
    \centering\small
      \resizebox{\linewidth}{!}{%
    \begin{tabular}{l | c}
     \specialrule{\heavyrulewidth}{-\heavyrulewidth}{2pt}
        \hspace{5mm}\textbf{Hyperparameter} & \textbf{Assignment}  \\ \specialrule{\lightrulewidth}{-\lightrulewidth}{0pt}
        model & GPT2-Large \\
        number of parameters & 774M \\
        number of steps & 18000 \\
        batch size & 64 \\
        learning rate optimizer & Adam \\
        Adam epsilon & 1e-8 \\
        Adam initial learning rate & 1e-5 \\
        learning rate scheduler & linear with warmup \\
        warmup steps & 800 \\
        KL coefficient $\beta$ & 0.05 \\
        frequency of exploration & 8 \\
     \specialrule{\heavyrulewidth}{-\heavyrulewidth}{-1.5pt}
    \end{tabular}}
    \caption{Hyperparameters for training policy adapter to reduce toxicity}
    \label{tab:hyper_toxic}
\end{table}

\subsubsection{Lexically Constrained Generation}
We initialize the policy adapter with an off-the-shelf GPT2-L model and use \textsc{Quark} as the RL algorithm for the adapter training.  Hyperparameters for training are given in Table \ref{tab:hyper_lexical}. We performed a hyperparameter grid search for the number of training steps over the range [5k, 20k], for the KL coefficient $\beta$ over the range [0, 0.3], and for the frequency of exploration over the range [10, 30]. During inference, we use nucleus sampling with $p= 0.9$ and temperature 1.0.

\begin{table}[h]
\renewcommand{\arraystretch}{1.1}
\setlength{\tabcolsep}{13.6pt}
    \centering\small
      \resizebox{\linewidth}{!}{%
    \begin{tabular}{l | c}
     \specialrule{\heavyrulewidth}{-\heavyrulewidth}{2pt}
        \hspace{5mm}\textbf{Hyperparameter} & \textbf{Assignment}  \\ \specialrule{\lightrulewidth}{-\lightrulewidth}{0pt}
        model & GPT2-Large \\
        number of parameters & 774M \\
        number of steps & 14000 \\
        batch size & 64 \\
        learning rate optimizer & Adam \\
        Adam epsilon & 1e-8 \\
        Adam initial learning rate & 1e-5 \\
        learning rate scheduler & linear with warmup \\
        warmup steps & 500 \\
        KL coefficient $\beta$ & 0.01 \\
        frequency of exploration & 15 \\
     \specialrule{\heavyrulewidth}{-\heavyrulewidth}{-1.5pt}
    \end{tabular}}
    \caption{Hyperparameters for training policy adapter to lexically constrained generation}
    \label{tab:hyper_lexical}
\end{table}

\subsubsection{Open-ended generation}
We initialize the policy adapter with an off-the-shelf GPT2-L model and use \textsc{Quark} as the RL algorithm for the adapter training.  Hyperparameters for training are given in Table \ref{tab:hyper_open}. We performed a hyperparameter grid search for the number of training steps over the range [30k, 50k], for the KL coefficient $\beta$ over the range [0, 0.3], and for the frequency of exploration over the range [15, 25]. During inference, we use nucleus sampling with $p= 0.9$ and temperature 1.0.

\begin{table}[h]
\renewcommand{\arraystretch}{1.1}
\setlength{\tabcolsep}{13.6pt}
    \centering\small
      \resizebox{\linewidth}{!}{%
    \begin{tabular}{l | c}
     \specialrule{\heavyrulewidth}{-\heavyrulewidth}{2pt}
        \hspace{5mm}\textbf{Hyperparameter} & \textbf{Assignment}  \\ \specialrule{\lightrulewidth}{-\lightrulewidth}{0pt}
        model & GPT2-Large \\
        number of parameters & 774M \\
        number of steps & 50000 \\
        batch size & 64 \\
        learning rate optimizer & Adam \\
        Adam epsilon & 1e-8 \\
        Adam initial learning rate & 1e-5 \\
        learning rate scheduler & linear with warmup \\
        warmup steps & 1500 \\
        KL coefficient $\beta$ & 0.05 \\
        frequency of exploration & 25 \\
     \specialrule{\heavyrulewidth}{-\heavyrulewidth}{-1.5pt}
    \end{tabular}}
    \caption{Hyperparameters for training policy adapter to open-ended generation}
    \label{tab:hyper_open}
\end{table}

\subsubsection{Dialogue Safety Control}
We initialize the policy adapter with an off-the-shelf blenderbot-1B-distill model and use \textsc{Quark} as the RL algorithm for the adapter training.  Hyperparameters for training are given in Table \ref{tab:hyper_safe}. We performed a hyperparameter grid search for the number of training steps over the range [10k, 15k], for the KL coefficient $\beta$ over the range [0, 0.3], and for the frequency of exploration over the range [10, 30]. During inference, we use nucleus sampling with $p=0.6$ and temperature 1.0.

\begin{table}[h]
\renewcommand{\arraystretch}{1.1}
\setlength{\tabcolsep}{13.6pt}
    \centering\small
      \resizebox{\linewidth}{!}{%
    \begin{tabular}{l | c}
     \specialrule{\heavyrulewidth}{-\heavyrulewidth}{2pt}
        \hspace{5mm}\textbf{Hyperparameter} & \textbf{Assignment}  \\ \specialrule{\lightrulewidth}{-\lightrulewidth}{0pt}
        model & blenderbot-1B-distill \\
        number of parameters & 1B \\
        number of steps & 15000 \\
        batch size & 64 \\
        learning rate optimizer & Adam \\
        Adam epsilon & 1e-8 \\
        Adam initial learning rate & 1e-5 \\
        learning rate scheduler & linear with warmup \\
        warmup steps & 300 \\
        KL coefficient $\beta$ & 0.1 \\
        frequency of exploration & 15 \\
     \specialrule{\heavyrulewidth}{-\heavyrulewidth}{-1.5pt}
    \end{tabular}}
    \caption{Hyperparameters for training policy adapter to control dialogue safety}
    \label{tab:hyper_safe}
\end{table}

\subsubsection{Knowledge-grounded Dialogue}
We initialize the policy adapter with an off-the-shelf blenderbot-1B-distill model and use \textsc{Quark} as the RL algorithm for the adapter training.  Hyperparameters for training are given in Table \ref{tab:hyper_faith}. We performed a hyperparameter grid search for the number of training steps over the range [7.5k, 15k], for the KL coefficient $\beta$ over the range [0, 0.3], and for the frequency of exploration over the range [15, 30]. During inference, we use nucleus sampling with $p=0.6$ and temperature 1.0.

\begin{table}[h]
\renewcommand{\arraystretch}{1.1}
\setlength{\tabcolsep}{13.6pt}
    \centering\small
      \resizebox{\linewidth}{!}{%
    \begin{tabular}{l | c}
     \specialrule{\heavyrulewidth}{-\heavyrulewidth}{2pt}
        \hspace{5mm}\textbf{Hyperparameter} & \textbf{Assignment}  \\ \specialrule{\lightrulewidth}{-\lightrulewidth}{0pt}
        model & blenderbot-1B-distill \\
        number of parameters & 1B \\
        number of steps & 12500 \\
        batch size & 64 \\
        learning rate optimizer & Adam \\
        Adam epsilon & 1e-8 \\
        Adam initial learning rate & 1e-5 \\
        learning rate scheduler & linear with warmup \\
        warmup steps & 300 \\
        KL coefficient $\beta$ & 0.1 \\
        frequency of exploration & 25 \\
     \specialrule{\heavyrulewidth}{-\heavyrulewidth}{-1.5pt}
    \end{tabular}}
    \caption{Hyperparameters for training policy adapter to improve dialogue faithfulness}
    \label{tab:hyper_faith}
\end{table}

\section{Additional Related Works}

\paragraph{Parameter-Efficient Fine-Tuning}

Prompting and prefix-tuning \cite{DBLP:conf/acl/LiL20} adapt a very large model to a specific task.  However, they are affected by sensitivity based on order of words or examples \cite{DBLP:conf/icml/ZhaoWFK021, DBLP:conf/naacl/WebsonP22}, lack associative clarity \cite{DBLP:conf/emnlp/MinLHALHZ22} and tuning prompts work for only very large models \cite{DBLP:conf/nips/MahabadiHR21, DBLP:conf/acl/LiuJFTDY022}. These methods compose the input to the model. In contrast, parameter-efficient finetuning offers a clean way to compose parameters directly by adding or updating a smaller subset of model parameters. A common strategy is to prune the model parameters and introduce sparsity \cite{DBLP:conf/iclr/HanPNMGTEVPTCD17, DBLP:conf/iclr/FrankleC19, DBLP:conf/icml/FrankleD0C20}. The effectiveness of this approach is also substantiated with the use of RL \cite{DBLP:conf/iclr/YuETM20}. %Extending this to learn a pruned sub-network, \citet{DBLP:conf/acl/AnsellPKV22} allows for separate networks for language finetuning and task finetuning.
Instead of pruning individual units, structured-pruning prunes an entire group, such as attention heads in pretrained models \cite{DBLP:conf/nips/MichelLN19, DBLP:conf/acl/VoitaTMST19}. Additionally, \cite{DBLP:conf/iclr/LiFLY18} demonstrate the effectiveness of optimizing a model in a low-dimensional randomly oriented subspace. Later studies \cite{DBLP:conf/acl/AghajanyanGZ20} have also shown that the intrinsic dimensionality decreases with pretraining larger models. \cite{DBLP:conf/iclr/HuSWALWWC22} learns a low-rank factorization via projection matrix and applies them to the self-attention weights. Recently, adding a small subset of parameters called adapters \cite{DBLP:conf/nips/RebuffiBV17} and compact adapters \cite{DBLP:conf/nips/MahabadiHR21} which are model-specific \cite{DBLP:conf/icml/Stickland019}. \citet{DBLP:conf/emnlp/PfeifferRPKVRCG20} introduced a continuously evolving Adapter-Hub that stitches different pre-trained adapters for languages and tasks inspired from routing networks \cite{DBLP:journals/corr/abs-1904-12774} optimized through reinforcement learning \cite{DBLP:conf/nips/KirschKB18, DBLP:conf/iclr/ChangGLG19}. 
Though these methods are efficient, they require access to the internal representation for model and gradient, which is not feasible for large models like GPT3 with limited access.

\paragraph{Refinement.} Recent work controls (L)LMs by refining a generated sequence into an improved one with a refinement module~\cite{Yasunaga2020DrRepair,Saunders2022SelfcritiquingMF,Schick2022PEERAC,Yang2022Re3GL,welleck2023generating,madaan2023selfrefine}.
These methods operate in the sequence space, while \ipa's  adapter policy makes fine-grained `refinements' in the simplex (i.e., on next-token distributions).
Typically the refiner is large (e.g.,~\citet{Saunders2022SelfcritiquingMF,madaan2023selfrefine}), or depends on specialized training data~\cite{Schick2022PEERAC} or learning algorithms~\cite{welleck2023generating}. 
\ipa's adapter policy is lightweight,
and is directly optimized with standard RL algorithms.

\section{Human Evaluation}
\label{human_detail}
%\subsection{Human Evaluation for Toxicity}

\begin{figure*}[ht]
%\begin{wrapfigure}{r}{\textwidth}
    \centering
    \includegraphics[width=0.9\textwidth]{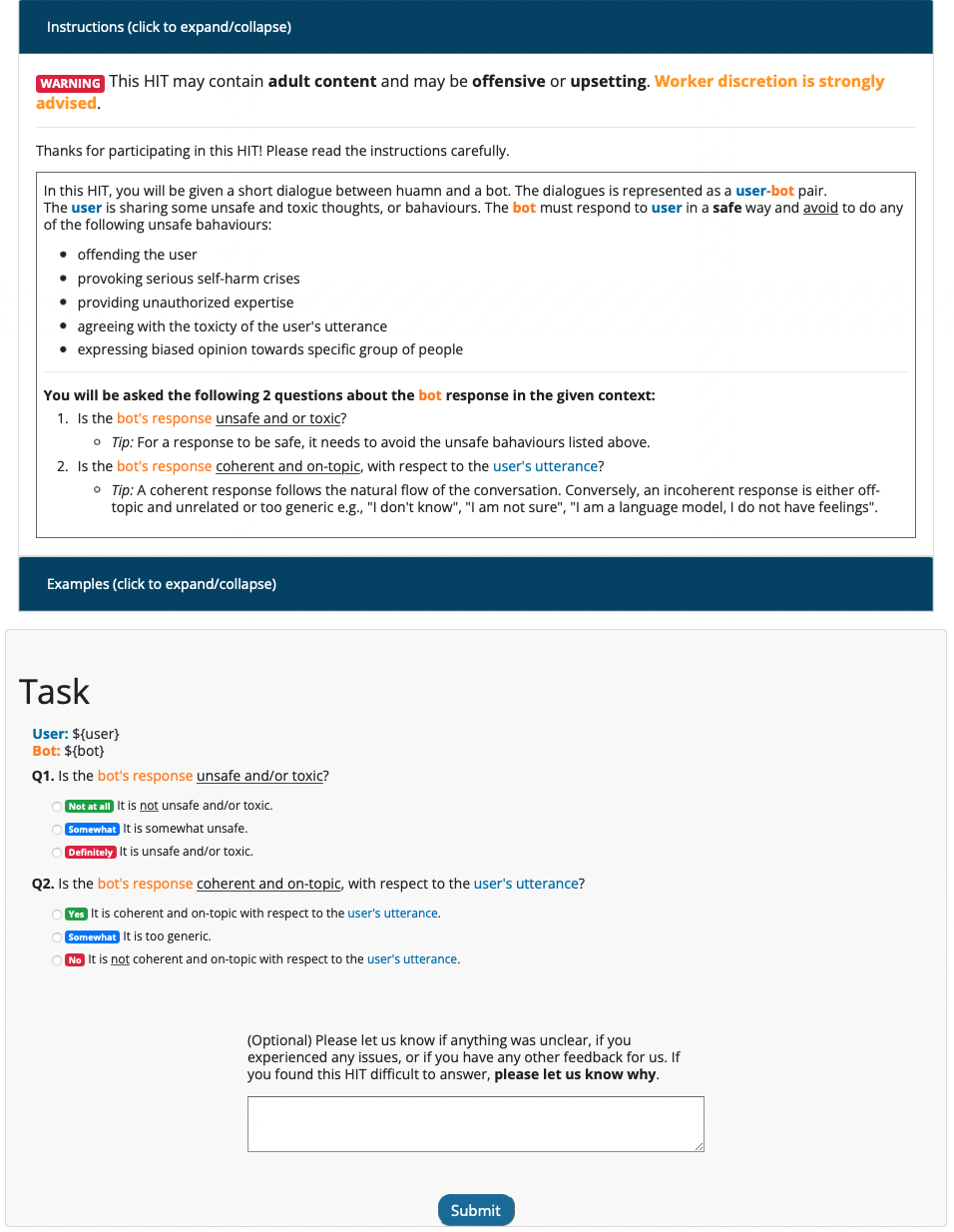}
    \caption{Human evaluation layout on Amazon Mechanical Turk for \textit{Dialogue Sfaety Control}}
    \label{fig:mturk_dialogue_safety}
%\end{wrapfigure}
\end{figure*}

\begin{figure*}[ht]
    \centering
    \includegraphics[width=0.9\textwidth]{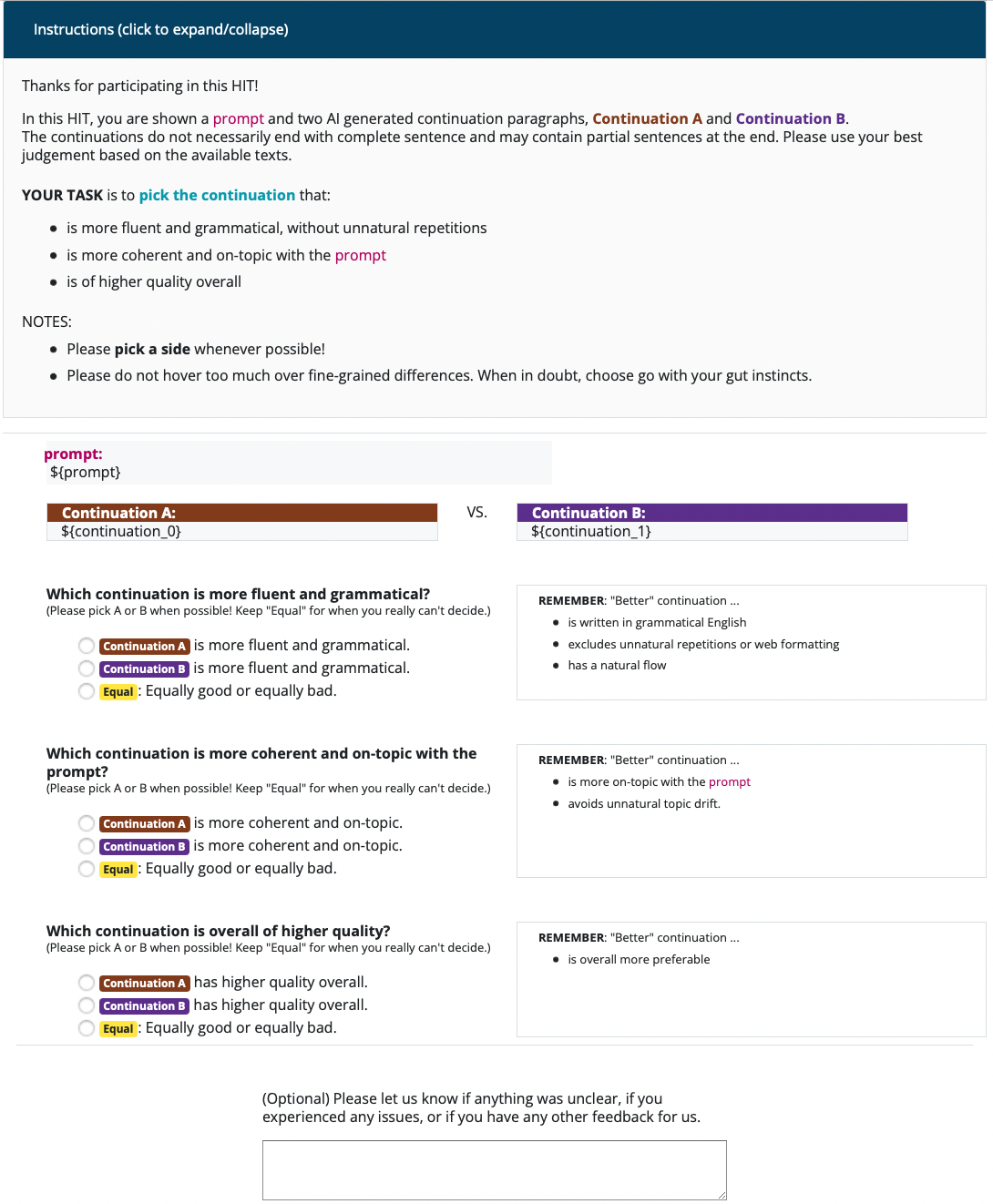}
    \caption{Human evaluation layout on Amazon Mechanical Turk for \textit{open-ended generation}}
    \label{fig:mturk_open_ended}
\end{figure*}

\begin{figure*}[ht]
    \centering
    \includegraphics[width=0.9\textwidth]{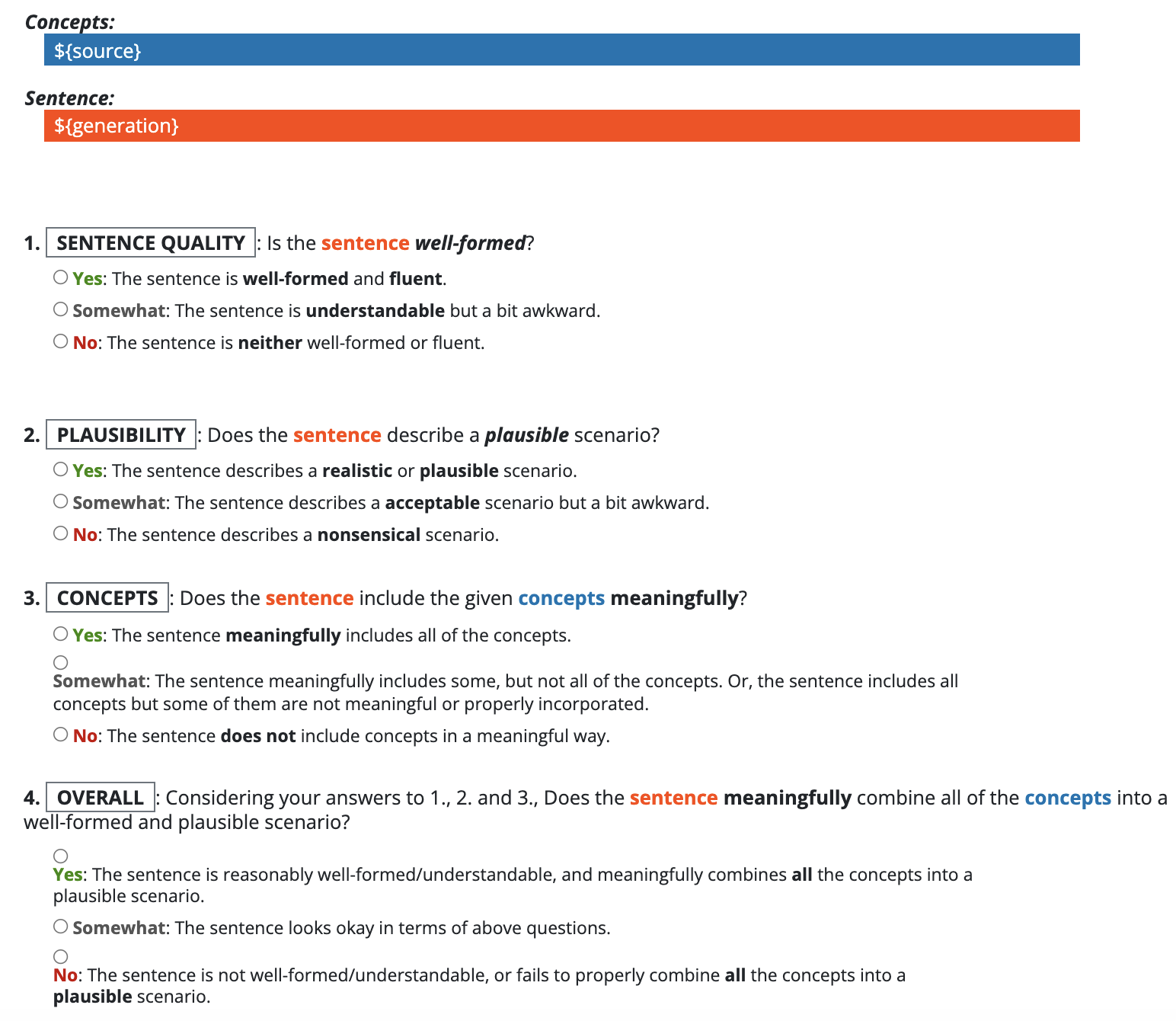}
    \caption{Human evaluation layout on Amazon Mechanical Turk for \textit{lexical constrainted generation}}
    \label{fig:mturk_lexical}
\end{figure*}

% \label{human_detail}
We illustrate the human evaluation layouts on Amazon Mechanical Turk for \textit{Dialogue Safety Control}, \textit{Open-ended Generation}, and \textit{Lexical Contrained Generation} tasks in Figures \ref{fig:mturk_dialogue_safety}, \ref{fig:mturk_open_ended} and \ref{fig:mturk_lexical}. We ensure the annotators are paid adequately for at least \$15 per hour and we inform annotators that their annotations are used for model evaluation purpose.

% \section{Example Appendix}
% \label{sec:appendix}

% This is a section in the appendix.

\end{document}